\pdfoutput=1
\documentclass[review]{elsarticle}

\usepackage{hyperref}
\usepackage{import}
\usepackage[export]{adjustbox}
\usepackage{amsfonts}
\usepackage{amsmath}
\usepackage{amssymb,amsmath,latexsym}
\usepackage{leftidx}
\usepackage{mathrsfs}
\usepackage{float}
\usepackage{chngpage}
\usepackage{orcidlink}
\usepackage{makecell}
\hypersetup{hidelinks} 

\usepackage{subcaption}
\usepackage{amsfonts}
\usepackage{amsmath}
\usepackage{amssymb,amsmath,latexsym}
\usepackage{leftidx}
\usepackage{mathrsfs}
\usepackage{floatflt}
\usepackage{bm}
\usepackage{longtable}
\usepackage{float}
\usepackage[figuresright]{rotating}
\usepackage{multirow}
\usepackage{colortbl}
\usepackage{booktabs}
\usepackage{diagbox}
\usepackage{caption}
\usepackage{graphicx}
\usepackage{caption}
\usepackage{multirow}
\usepackage{xurl}
\usepackage{xr}

\makeatletter
\newcommand*{\addFileDependency}[1]{
\typeout{(#1)}
\@addtofilelist{#1}
\IfFileExists{#1}{}{\typeout{No file #1.}}
}\makeatother

\usepackage{enumerate}
\usepackage{cases}
\usepackage{geometry}
\definecolor{tabcolor}{rgb}{.105,.410,.113}

\allowdisplaybreaks[4]

\usepackage{chngcntr}

\makeatletter
\def\ps@pprintTitle{%
  \let\@oddhead\@empty
  \let\@evenhead\@empty
  \let\@oddfoot\@empty
  \let\@evenfoot\@oddfoot
}
\makeatother

\begin{document}
\begin{frontmatter}

\title{Early detection of disease outbreaks and non-outbreaks using incidence data}
\author[1]{Shan Gao\orcidlink{0009-0002-1705-8033}}
  \ead{sgao9@ualberta.ca}
  \cortext[cor1]{Corresponding author.}
\author[1]{Amit K. Chakraborty\orcidlink{0000-0002-7960-3984}}
\author[2,3]{Russell Greiner}
\author[4,5,6]{Mark A. Lewis\orcidlink{0000-0002-7155-7426}}
\author[1,6]{Hao Wang\orcidlink{0000-0002-4132-6109}}
  
\address[1]{Department of Mathematical and Statistical Sciences, University of Alberta, AB, Canada}
\address[2]{Department of Computing Science, University of Alberta, AB, Canada}
\address[3]{Alberta Machine Intelligence Institute, AB, Canada}
\address[4]{Department of Mathematics and Statistics, University of Victoria, BC, Canada}
\address[5]{Department of Biology, University of Victoria, BC, Canada}
\address[6]{These authors jointly supervised this work.}

\date{}

\begin{abstract}

Forecasting the occurrence and absence of novel disease outbreaks is essential for disease management. Here, we develop a general model, with no real-world training data, that accurately forecasts outbreaks and non-outbreaks. We propose a novel framework, using a feature-based time series classification method to forecast outbreaks and non-outbreaks. We tested our methods on synthetic data from a Susceptible–Infected–Recovered model for slowly changing, noisy disease dynamics. Outbreak sequences give a transcritical bifurcation within a specified future time window, whereas non-outbreak (null bifurcation) sequences do not. We identified incipient differences in time series of infectives leading to future outbreaks and non-outbreaks. These differences are reflected in 22 statistical features and 5 early warning signal indicators. Classifier performance, given by the area under the receiver-operating curve, ranged from $0.99$ for large expanding windows of training data to $0.7$ for small rolling windows. Real-world performances of classifiers were tested on two empirical datasets, COVID-19 data from Singapore and SARS data from Hong Kong, with two classifiers exhibiting high accuracy. In summary, we showed that there are statistical features that distinguish outbreak and non-outbreak sequences long before outbreaks occur. We could detect these differences in synthetic and real-world data sets, well before potential outbreaks occur.\\

\end{abstract}

\begin{keyword}
infectious disease, early warning signal, early detection, time series classification, Susceptible-Infected-Recovered (SIR) model
\end{keyword}
\end{frontmatter}

\section{Introduction}
The occurrence and recurrence of infectious diseases are prevalent worldwide, with some infectious diseases showing a high mortality rate and/or strong transmissibility, while others may be nonfatal and disappear quickly without attracting much attention~\cite{jones2020history}. Balancing the risks associated with new infectious diseases and the costs of preventative measures, along with their consequences, has been a prolonged challenge. In the case of recurrent diseases like influenza, prior experience allows for informed responses, whereas we have less guidance and certainty with novel diseases. One recent example is the initial response to COVID-19, where governments started with implementing less restrictive measures, such as recommending face masks in public places, rather than enforcing mandatory quarantine policies, to minimize negative impacts on livelihoods and the economy~\cite{anderson2020will, gargoum2021limiting, de2020initial}. Early preventative intervention is associated with lower incidence~\cite{tian2020investigation, evans2024effectiveness}. Anticipating infectious disease outbreaks and non-outbreaks, ideally in an early and accurate manner, therefore becomes imperative to prevent misjudgments of health risk perceptions faced by societies and their citizens, guiding the implementation of mitigation measures.

The disease outbreaks, where incidence counts escalate rapidly from a negligible level (pre-pandemic) to an uncontrollable state (pandemic) following subtle disturbances, can be viewed as critical transitions. Many natural and societal systems exhibit the potential of critical transitions, experiencing abrupt shifts from one stable state to another~\cite{may1977thresholds, scheffer2020critical, scheffer2001catastrophic, kefi2007spatial}. Predicting disease outbreaks (non-outbreaks) can be reframed as predicting (no) critical transition.

Many efforts have been invested in these critical transition (that is, disease outbreak) forecasts through mathematical epidemiology modeling. These models can reflect the dynamics of infectious diseases, predict the outbreak occurrence (via the basic reproduction number $R_0$, a dimensionless value representing the expected number of secondary infections caused by a single infectious individual in a completely susceptible population~\cite{heffernan2005perspectives}), estimate the long-term severity of the pandemic (using final size, representing the proportion of individuals who ultimately become infected over the disease transmission process), and hence significantly help in disease prevention and control measures. However, understanding the disease transmission mechanisms and formulating context-specific mathematical models often necessitate sufficient data for model parameterization~\cite{metcalf2017opportunities}, which contradicts the limited availability of collected data at the early stage. Disease control and prevention are also time-critical endeavors. While delving into the intricate mechanisms can be data-oriented and time-consuming, prompt mitigation practices during the initial stage of a pandemic are more urgent. Mathematical modeling is better suited for mid-pandemic research and post-pandemic reflection rather than providing immediate guidance in the initial stages.

Alternatively, the early warning signal (EWS) emerges as a promising model-independent method for the early prediction of critical transitions. The fundamental principle behind the anticipation is the critical slowing down (CSD) phenomenon: a decrease in local stability when the system approaches a critical threshold, resulting in sudden shifts of alternate stable states following small perturbations~\cite{van2007slow}. Various indicators, such as increasing standard deviation~\cite{carpenter2006rising}, growing coefficient of variance~\cite{carpenter2006rising}, rising autocorrelation at lag-1~\cite{dakos2010spatial}, changing values of skewness~\cite{guttal2008changing}, and escalating kurtosis~\cite{biggs2009turning}, can be used as early warning indicators (EWIs) to identify the presence of CSD. 

Despite the previous success of applying EWS for predicting forthcoming catastrophic transitions~\cite{tredennick2022anticipating, harris2020early, o2021early}, the reliability and applicability of EWS methods using the CSD as an indicator are still under debate~\cite{hastings2010regime, dakos2015resilience}. The inherent stochasticity, common in disease transmission and surveillance data, can result in unforeseen dynamics and undermine the critical transition prediction. Furthermore, the smoothness assumption in EWS models may not hold in many real-world scenarios, rendering the classic “ball-in-cup” metaphor inapplicable~\cite{hastings2010regime}. This lack of smoothness is particularly evident in disease transmission dynamics (the pathogen evolution can be abrupt, for example). To obtain more robust and accurate predictions, we apply machine learning techniques, which have been proven effective in predicting transitions by training upon simulations~\cite{bury2021deep, kong2021machine, deb2022machine}. 

Current endeavors on EWS theory typically concentrate on a singular object featuring critical transitions. Many investigations focus on how the changing trends in solely one indicator (or the combination of specific features) of the time series before critical transitions can be exploited~\cite{carpenter2006rising, dakos2018identifying, boers2021observation}. In the realm of disease-related prevention and control, accurate prediction of non-outbreaks is also vital. In our context, a machine-learned classifier requires incidence time series data containing critical transitions (outbreaks) and null transitions (non-outbreaks) as the training set to capture the underlying patterns of each scenario. However, incidence data with non-outbreak are inherently scarce or usually incomplete because data collection of diseases with negligible consequences tends to be overlooked or discontinued prematurely. Considering these limitations in data availability and the need for more control over data, synthetic data can serve as an alternative option for the model training process.

We adopt the susceptible-infected-recovered (SIR) model~\cite{kermack1927contribution} as the basis for simulating disease transmission dynamics. The SIR model, comprising susceptible, infected, and recovered individuals, can roughly reflect disease transmission dynamics in the real world. In such a deterministic model where all parameters remain the same, the epidemiological system can reach two alternate statuses: disease-free equilibrium $E_1$ if $R_0<1$ or disease outbreak equilibrium $E_2$ if $R_0>1$. Additionally, the dynamics in the real world are complex and coupled with exogenous and endogenous noise and stochasticity, unlike the deterministic model. To account for this, we introduce three types of noise—white noise, multiplicative environmental noise, and demographic noise—to create three noise-induced SIR models for data simulation.

\begin{figure}[!ht]
    \centering
    \includegraphics[width=\textwidth]{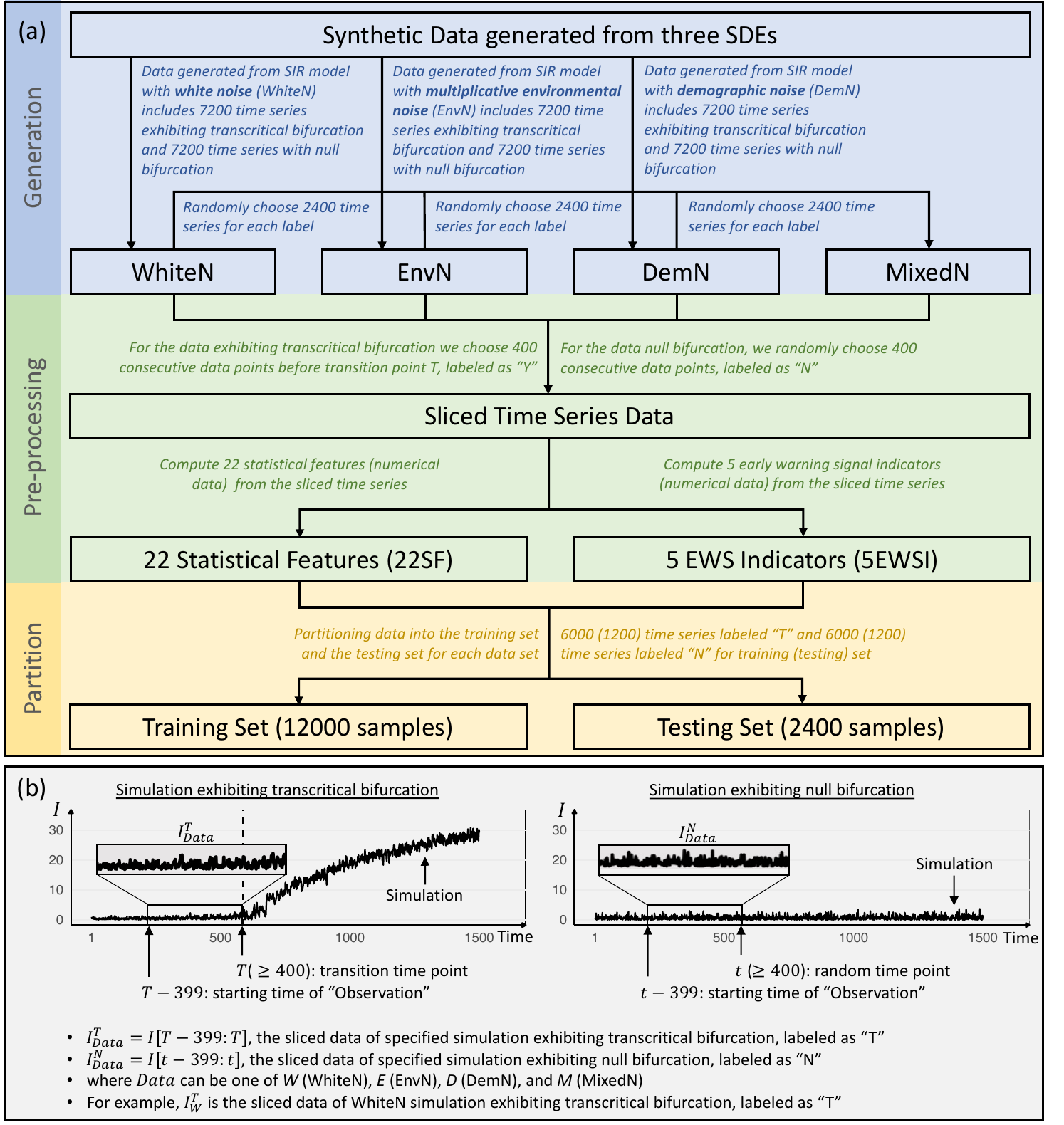}
    \caption{Overview of the synthetic data utilized in the study. (a) The blue frame represents the data generation step, the green frame signifies data pre-processing, and the yellow frame indicates data partitioning. (b) Examples of simulation from the SIR model with white noise.}
    \label{SummarySynthetic}
\end{figure}

We simulate data exhibiting critical transition by slowly increasing the transmission rate $\beta$ and fixing other parameters. Subsequently, $R_0(t)$ becomes time-dependent and monotonically increasing. We generate 14,400 replicates of time series $I(t)$ ($t=1,2,...,1500$) from each noise-induced SIR model, with half exhibiting a transcritical bifurcation (where $R_0(t)$ increasing through $1$ for some $T\in[401,1500]$) and the other half a null bifurcation (where $R_0(t)<1$ for any $t\in[1,1500]$). Finally, we define the subsequence of replicate exhibiting transcritical bifurcation $I[1:T]$ as the pre-transition data. We label the subsequence $I[T-399:T]$ as “T” and label the subsequence of null bifurcation data $I[t-399:t]$ for a random $t\in[400,1500]$ as “N” (see Figure~\ref{SummarySynthetic} (b)). The parameterization is justified in Chakraborty. et al. \cite{chakraborty2024early}. Data simulation and labeling details are further illustrated in Methods~\ref{DataGeneration}.

After obtaining the simulation datasets, we train the classifier using the following features, computed from $I[1:T]$ and $I[t-399:t]$: (1) 22 statistical features (22SF, see Supplementary Table~\ref{22F}) for their outstanding performance on time series classification (TSC) tasks and minimal redundancy~\cite{lubba2019catch22}, and (2) 5 early warning signal indicators (5EWSI, see Supplementary Table~\ref{5S}) to represent time series data. We considered the following machine learning algorithms: (1) the gradient boosting machine (GBM), (2) the logistic regression model (LRM), (3) k-nearest neighbor (KNN), and (4) support vector machines (SVM).

Here, instead of forecasting the incidence counts or the death rate of the infectious disease, our focus lies on the early prediction of outbreak versus non-outbreak. We construct a framework for predicting pandemics using incomplete time series (i.e. $I[T-399:T]$ and $I[t-399:t]$) to address the challenges inherent in disease control and prevention. We train a set of classifiers using data simulated from noise-induced SIR models. We then test the classifiers on withheld simulated data to assess the performance. To investigate the adaptability, we vary the input lengths or the gap between the input and transition point (these two methods are referred to as the Rolling window and Expanding window, respectively; see Figures~\ref{RollingAUC}-\ref{ExapndingAUC}) and conduct the training-testing process. In the end, we test the classifiers on real-world COVID-19 incidence data that undergo transcritical bifurcation (outbreaks) and SARS data that exhibits no bifurcation (non-outbreak) ever since a certain time point. Our work establishes a link between machine learning practices informed by mathematical modeling data and critical transition prediction for disease prevention and control, two disciplines that have rarely interacted cohesively before.

\section{Results}
\begin{figure}[!ht]
    \centering
    \includegraphics[width=\textwidth]{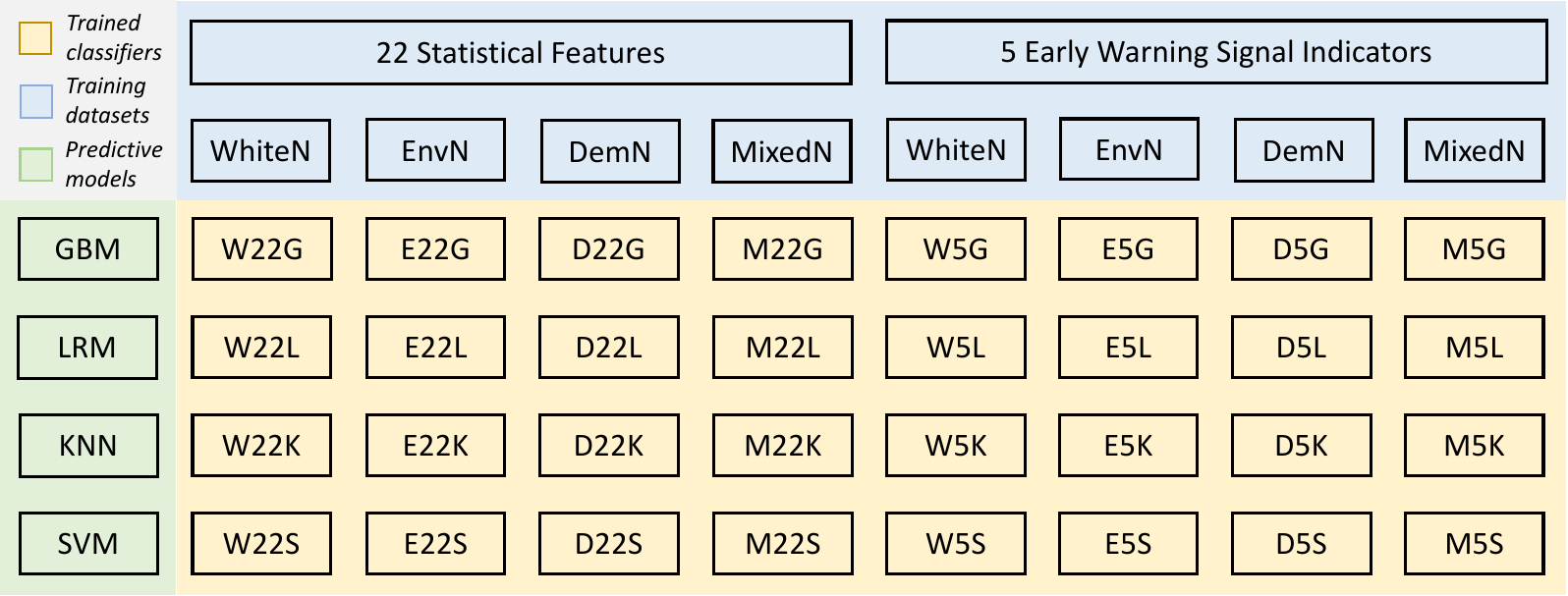}
    \caption{32 classifiers trained from every unique combination of four synthetic training datasets, two feature extraction libraries, and four predictive models. The initial letter in each classifier's name corresponds to the training set (W, WhiteN; E, EnvN; D, DemN; M, MixedN). The middle number represents the feature extraction (22, 22 statistical features; 5, 5 early warning signal indicators). The last letter represents the predictive model (G, GBM; L, LRM; K, KNN; S, SVM).}
    \label{trainedmodels}
\end{figure}

We train $4\times2\times4=32$ classifiers, denoted as W22G-M5S (Figure~\ref{trainedmodels}), based on every unique combination of four simulated data sets, two feature extraction libraries, and four predictive models (see Methods~\ref{models}). To illustrate our findings, we assess the performance of these classifiers on withheld synthetic testing sets using accuracy and the area under the receiver-operating curve (AUC) metrics.

Henceforth, we use the notation WhiteN (EnvN, DemN, or MixedN) to denote datasets comprising 14,400 replicates simulated from the SIR model with white noise (multiplicative environmental noise, demographic noise, or a mixture from other three datasets) in the following content. Sequences labeled as “T” (resp., “N”) represent pre-transition subsequences of time series experiencing transcritical bifurcation – i.e., eventual outbreak (resp., subsequences exhibiting null bifurcation –  i.e., normality). For simplicity, $I_{Data}$, where $Data$ is one of $W$ (WhiteN), $E$ (EnvN), $D$ (DemN), or $M$ (MixedN), represents the sliced data replicates of specified simulation. $I_{Data}^{Label}$ with $Label$ as $T$ (outbreak) or $N$ (non-outbreak), represents labeled $I_{Data}$. Additionally, 22SF (5EWSI) of $I_{Data}^{Label}$ refers to 22SF (5EWSI) computed from the specified replicates. 

\subsection{Mann-Whitney U test results on synthetic data}
Features extracted from sequences with each label exhibit non-normal distribution (see Supplementary Figures~\ref{boxW}-\ref{boxM} for 22SF and Supplementary Figures~\ref{SboxW}-\ref{SboxM} for 5EWSI). Therefore, we conduct the Mann-Whitney U test, which is non-parametric and does not assume specific data distribution, on 22SF and 5EWSI computed from “T” and “N” replicates. We find statistically significant differences between features derived from “T” sequences and those from “N” sequences with $p\ll0.001$. These findings indicate the potential to statistically discriminate between these two categories based on 22SF or 5EWSI.

\subsection{Performance on withheld testing set}
We evaluate the classification performance of synthetic-data-trained classifiers on withheld testing sets, as presented in Figure~\ref{FullAccuracy_withheld} (detailed data are in Supplementary Table~\ref{SythResult}). Presumably, due to relatively easy classification tasks, results suggest that all the classifiers can achieve near-perfect performance on withheld testing sets, with AUC scores ranging from E5K ($0.9911\pm0.0038$) to D22G, D22K, D22L, D22S, D5G, D5K, D5L, D5S, W22L, W22S, W5G, and W5L ($1\pm0.0000$). This finding aligns with our speculation based on the Mann-Whitney U test results.

The AUC scores of 32 different classifiers are consistently high, exceeding $0.9900$ and being numerically close. Hence we conduct DeLong tests~\cite{delong1988comparing} to compare the AUC scores of different classifiers. Supplementary Tables~\ref{DeLong_GBM}-\ref{DeLong_SVM} present p-values of the DeLong tests across eight training sets with a predetermined model (GBM, LRM, KNN, and SVM, respectively). Classifiers trained using 5EWSI of $I_E$ (i.e., E5G, E5L, E5K, or E5S) generally exhibit slightly lower AUC scores compared to classifiers trained using the same predictive model but with different datasets ($p<0.001$ for most classifiers).

Supplementary Tables~\ref{DeLong_white_22SF}-\ref{DeLong_mixed_EWSI} present the DeLong test results among four models, trained on the same dataset. Counterintuitively, despite the distinct properties of various predictive models and their ability to handle different types of data, the test results suggest no significant difference in AUC scores among four predictive models when applied to most of the synthetic testing sets. However, there are some exceptions: when using 22SF of $I_E$ (see Supplementary Table~\ref{DeLong_env_22SF}), the AUC of four classifiers, E22K has the lowest AUC score $(0.9956\pm0.0027)$ compared to E22S ($0.9993\pm0.0011$, $p<0.01$), E22L ($0.9993\pm0.0011$, $p<0.01$), or E22G ($0.9991\pm0.0012$, $p<0.01$). Moreover, as illustrated in Figure~\ref{FullAccuracy_withheld}, classifiers trained by 5EWSI of $I_E$ across four models rank the last four in terms of AUC scores, with E5K ($0.9911\pm0.0038$, $p<0.05$) being the lowest, while the AUC scores among E5G, E5S, and E5L are statistically similar ($p>0.05$). This means 5EWSI of $I_E$ is the most challenging one to be distinguished among the four models.

\begin{figure}[!ht]
    \centering
    \includegraphics[width=\textwidth]{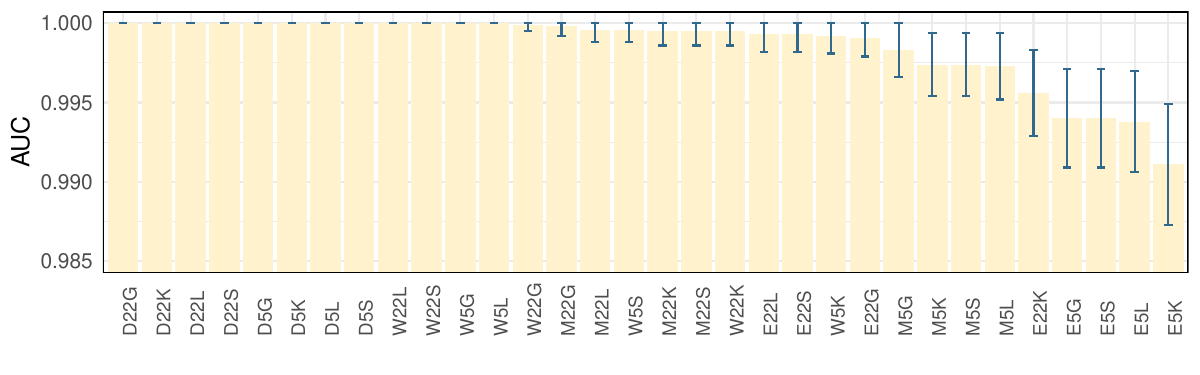}
    \caption{The AUC values of 32 synthetic-data-trained classifiers (horizontal axis, see Figure~\ref{trainedmodels}) on withheld testing sets. Classifiers are reordered by AUC scores. Error bars correspond to the $95\%$ confidence intervals. DeLong tests are conducted to compare the AUC values of classifiers, with detailed results available in Supplementary Tables~\ref{DeLong_GBM}-\ref{DeLong_SVM} (where the predictive model is fixed) and Supplementary Tables~\ref{DeLong_white_22SF}-\ref{DeLong_mixed_EWSI} (where the training data is fixed).}
    \label{FullAccuracy_withheld}
\end{figure}

\subsection{Performance on withheld testing set with varying sequence length and gap}
\begin{figure}[!ht]
    \centering
    \includegraphics[width=\textwidth]{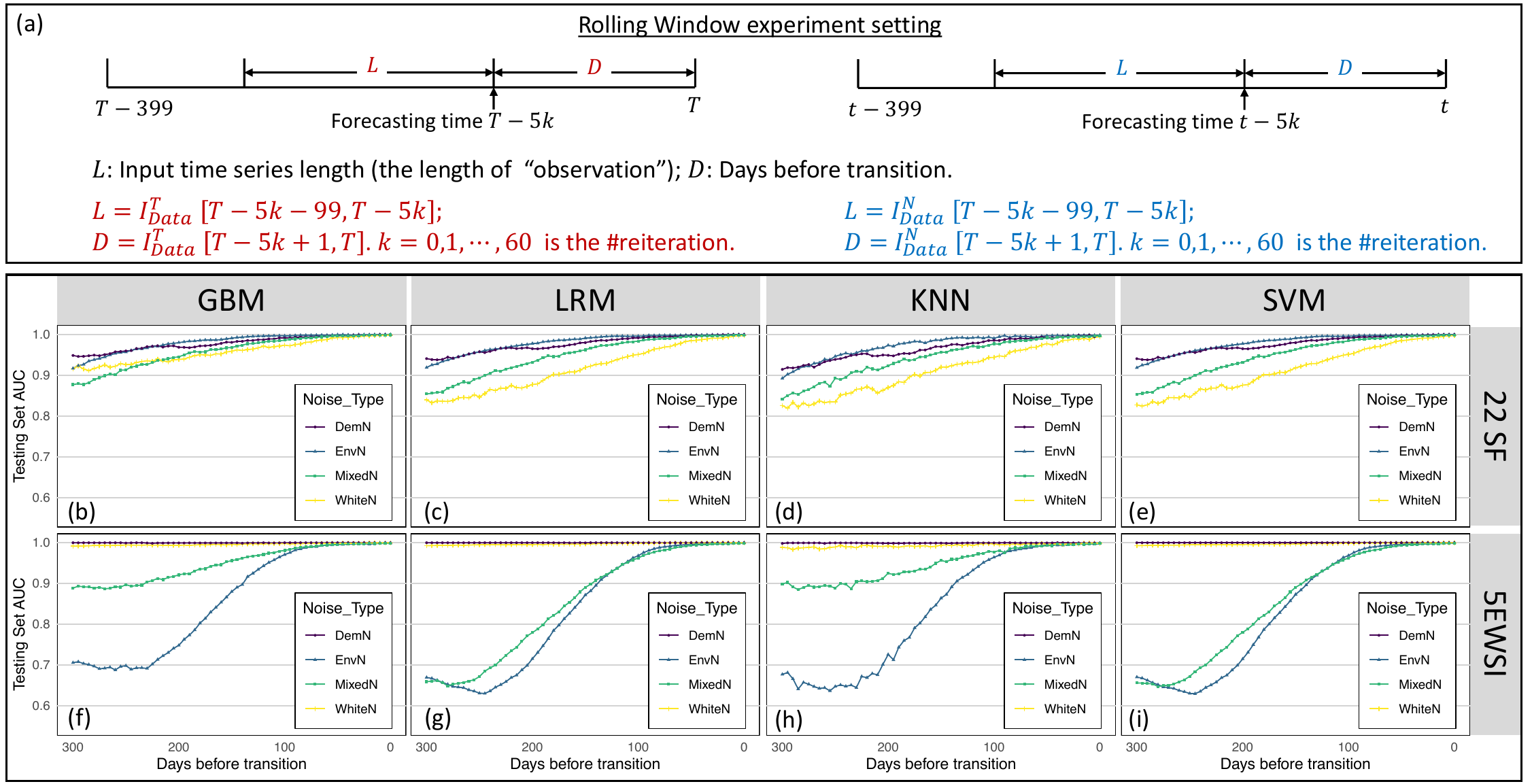}
    \caption{Illustration in Rolling window experiment settings and AUC score at different reiterations. Figure (a) exhibits the Rolling window settings on $I_{Data}^T$ and $I_{Data}^N$. We maintain a fixed length of input time series $L$ and roll the window $L$ from the right to the left boundary. Figures (b)-(i) are AUC scores, depicting the classification performance of the classifier trained from changing gaps $D$ between the subsequence and transition points. Two feature extraction methods (22SF, 22 statistical features; 5EWSI, 5 early warning signal indicators) are put in two rows and four predictive models (GBM, gradient boosting machine; LRM, logistic regression model; KNN, k-nearest neighbor; SVM, support vector machine) are presented in four columns.}
    \label{RollingAUC}
\end{figure}

\begin{figure}[!ht]
    \centering
    \includegraphics[width=\textwidth]{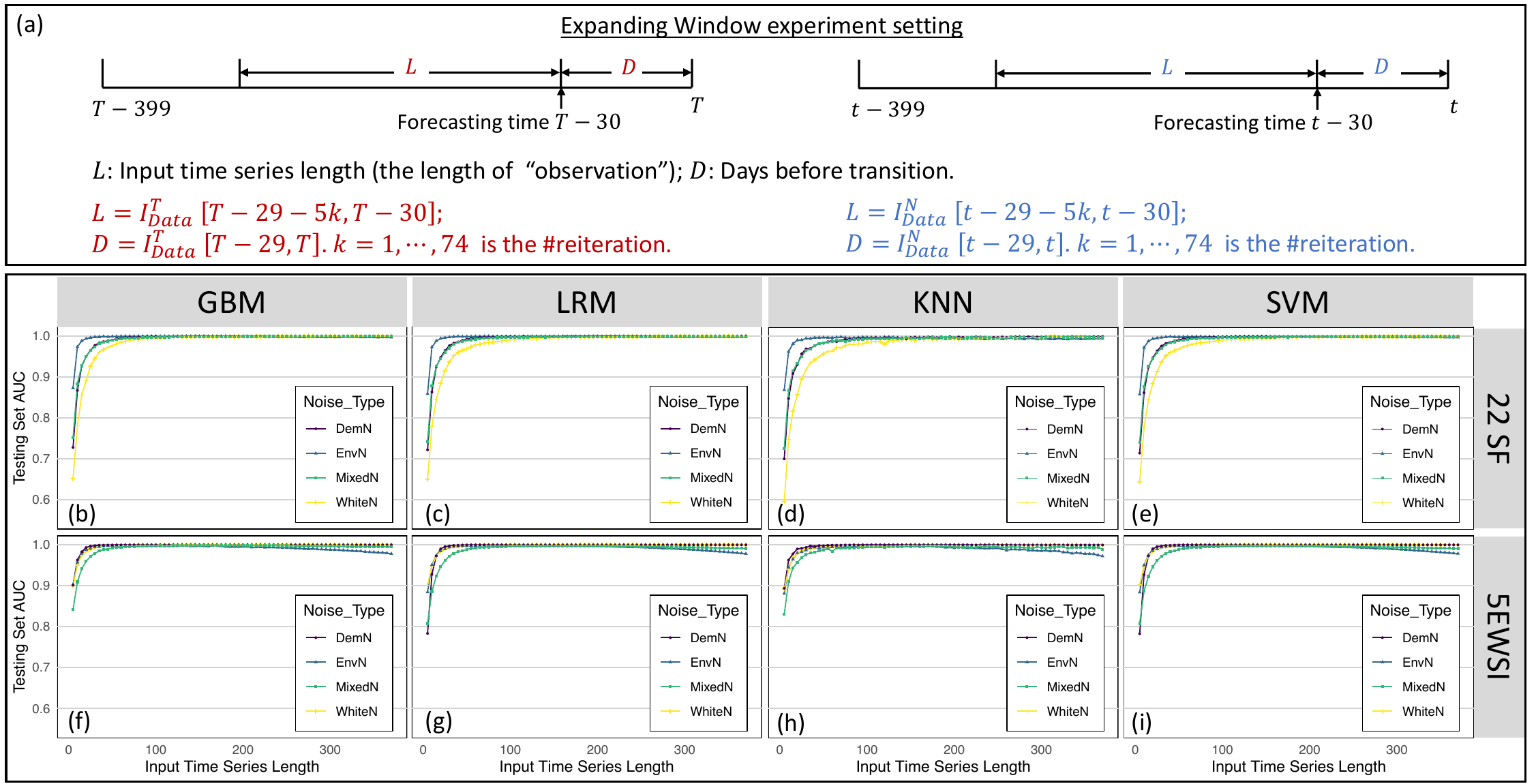}
    \caption{Illustration in Expanding window experiment settings and AUC score at different reiterations. Figure (a) exhibits the Expanding window settings on $I_{Data}^T$ and $I_{Data}^N$. We fix the length of $D$ as $30$, then expand $L$ until reaching the left boundary of sliced data. The initial length of $L$ is 5. Figures (b(-(i) are AUC scores, depicting the classification performance of the classifier trained from changing gaps $D$ between the subsequence and transition points. Two feature extraction methods (22SF, 22 statistical features; 5EWSI, 5 early warning signal indicators) are put in two rows and four predictive models (GBM, gradient boosting machine; LRM, logistic regression model; KNN, k-nearest neighbor; SVM, support vector machine) are presented in four columns.}
    \label{ExapndingAUC}
\end{figure}

In this context, replicates $I_{Data}^T$ end at the transition points, leaving no response time for implementing mitigation measures. Despite the satisfactory performance of $32$ classifiers over withheld testing sets, two questions naturally emerge: (1) How early can we predict an impending disease outbreak? (2) How much data is required for prediction to maintain a certain level of accuracy? To address these concerns, we conduct two experiments using the Rolling window (where the distance between data and the transition point varies while fixing the data length) and the Expanding window (where the data length varies while the distance between data and the transition point is fixed), as illustrated in Figure~\ref{RollingAUC} (a) and Figure~\ref{ExapndingAUC} (a). We repeatedly conduct training-testing processes using Rolling windows or Expanding windows, resulting in corresponding classifiers at each round. The AUC values and classification accuracy are depicted by points using different colors and shapes. Results are shown in Figures~\ref{RollingAUC}-\ref{ExapndingAUC} for AUC and Supplementary Figure~\ref{EarlinessACC} for accuracy.

The results of the Rolling window (Figure~\ref{RollingAUC} (b)-(i)) illustrate that approaching the transition point can generally increase the AUC scores for all classifiers, among which those trained from the GBM model showing better performance across four types of noise data. Classifiers trained on 5EWSI of any noise data can achieve AUC scores of almost $1$ when the gap is shorter than $50$, while those employing 22SF require a distance of less than $30$ to reach similar performance as 5EWSI. Classifiers trained on 22SF of $I_E$ and $I_D$ slightly outperform those trained on the other two noise data for all four models (see blue triangle and purple round points in Figure~\ref{RollingAUC} (b)-(e)). However, 5EWSI of $I_E$ and $I_M$ exhibit notably lower AUC across predictive models when the distance is greater than $100$ (see blue triangle and green square points in Figure~\ref{RollingAUC} (f)-(i)). We also observe that when the distance between the time series to the transition point is less than $100$, two feature extractions computed from any of the sliced time series exhibit similarly excellent performance across four models. As the window moves farther from the transition point, at a distance of $300$ data points for example, the GBM model can achieve better performance for both $I_D$ and $I_M$ in terms of feature extraction. We further notice that classifiers trained from 5EWSI of $I_W$ or $I_D$ can achieve better and more robust performance as their AUC values do not increase substantially along with closer distances (see yellow cross and purple round points in Figure~\ref{RollingAUC} (f)-(i)). In contrast, classifiers trained from 5EWSI of $I_D$ and $I_M$ yield significantly improved classification performance when far from the transition points (see green square and blue triangle points in Figure~\ref{RollingAUC} (f)-(i)). Additionally, it is worth noting that the results of LRM and SVM models present negligible differences (less than $0.001$ in general).

The results of the Expanding window experiments are presented in Figure~\ref{ExapndingAUC} (b)-(i). AUC scores exhibit a rapid improvement as we increase the length of the testing time series from $5$ to $100$, preserving an AUC score of over $0.97$ thereafter, which suggests that the classifiers derived from the framework can remain applicable even for short synthetic time series. While longer time series data intuitively contribute to better performance, we observe a slight decreasing trend in AUC for classifiers trained by 5EWSI when the input time series length approaches $370$, likely due to redundant data distant from the transition point. Interestingly, classifiers trained from 22SF of the original time series show no such decline in AUC. One speculation is that classifiers trained on data represented by 22SF are more robust when dealing with data redundancy.

\subsection{Performance on empirical testing set}
\label{Performance on Empirical Testing Data}
\begin{figure}[!ht]
    \centering
    \includegraphics[width=\textwidth]{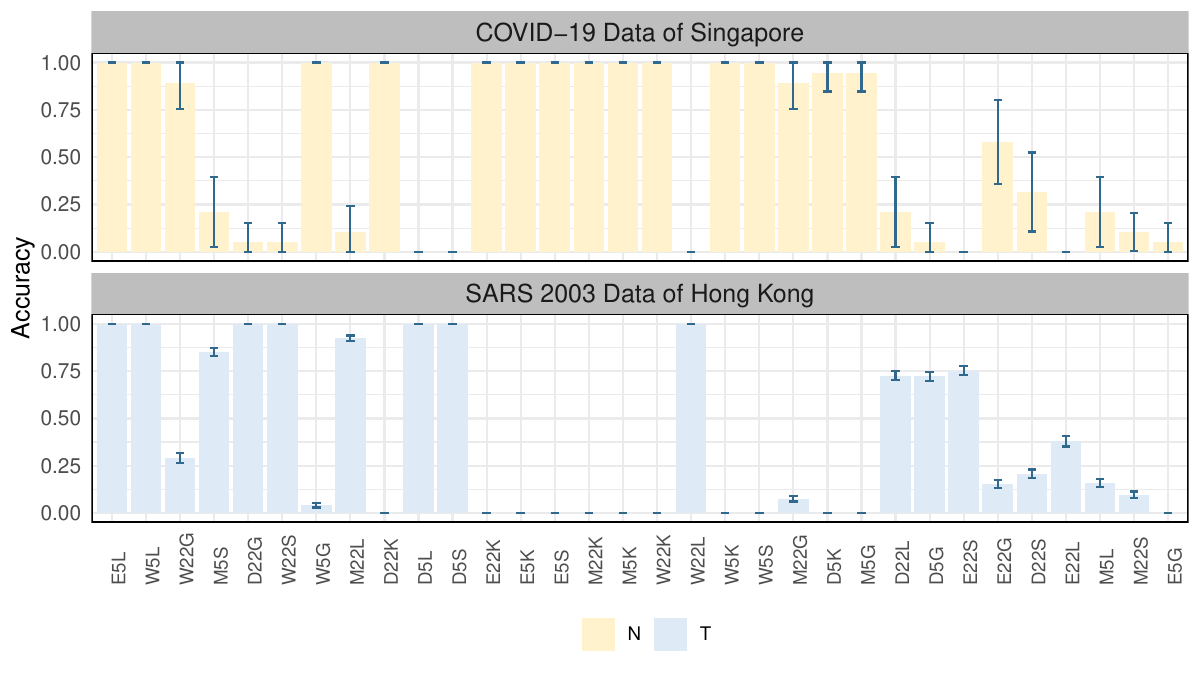}
    \caption{Classification accuracy of 32 synthetic-data-trained classifiers (horizontal axis, see Figure~\ref{trainedmodels}) to correctly distinguish pre-transition (“T”) and normality (“N”) subsequences of three empirical datasets. Error bars correspond to the $95\%$ confidence intervals.}
    \label{FullAccuracy}
\end{figure}

We test the performance of 32 trained classifiers (see Figure~\ref{trainedmodels}) on two real-world infection datasets: COVID-19 data from Singapore ($I_{SG}$) and SARS data from Hong Kong ($I_{HK}$). Although violating the assumption of homogeneous population distribution in the SIR model, we additionally utilize COVID-19 data from 18 countries as the third test set ($I_{18}$, comprising 194 replicates; see Supplementary section \ref{I18}) for two purposes: evaluating the performance of classifiers on data that deviates from the model assumption, and comparing classification performance with that of $I_{SG}$. Note that each dataset solely contains replicates labeled as either “T” or “N”, making it impossible to compute AUC scores. As a result, we only use the classification accuracy as the performance measurement on testing sets, and the results are presented in Figure~\ref{FullAccuracy}.

Classification results on $I_{SG}$ are presented in Figure~\ref{FullAccuracy}. We observe that classifiers trained from the KNN model outperform those trained by the other three models, among which $D5K$ has the lowest accuracy $0.0.9474$ $\pm 0.1004$, and the other classifiers have an accuracy of $1$ $\pm 0.0000$. Classifiers trained from 5EWSI of $I_W$ (i.e. W5G, W5L, W5K, and W5S) or $I_E$ (i.e. E5G, E5L, E5K, and E5S) exhibit consistent high classification accuracy. We observed much lower accuracy on $I_{18}$ compared to $I_{SG}$ (see Supplementary Table~\ref{18CovidResult}), which is consistent with our rationale that Singapore data is more suitable for SIR model assumptions.

While many classifiers can predict the potential outbreaks (that is, correctly classify replicates labeled as “T”) on both synthetic and empirical scenarios, we observe very different results on $I_{HK}$ that represent non-outbreaks (see Figure~\ref{FullAccuracy}). Remarkably, classifiers trained from KNN achieve an accuracy of more than 0.9 for most cases but fail in predicting “N” replicates. One presumable explanation for this is that the KNN model has the inherent tendency to classify any input data as “T” cases. Still, there are seven classifiers, namely E5L, W5L, D22G, W22S, D5L, D5S, and W22L, reach the accuracy of $1$ $(\pm0.0000)$. Additionally, many classifiers (W22L, W5L, E5L, D5L, M22L, and D22L) trained from LRM, the simplest one among the alternative four models, exhibit relatively better performance in general, except E22L ($0.3792$ $\pm 0.0275$) and M5L ($0.1575$ $\pm 0.0206$). 

The performance of COVID-19 and SARS data, or in other words, between replicates labeled as “T” (outbreaks) and “N” (non-outbreaks) seem to be complementary. Most classifiers performing well in classifying “T” replicates fail to predict replicates labeled as “N”, except for two classifiers, E5L and W5L, achieving an accuracy of $1$ $(\pm0.0000)$ in out-of-sample empirical data classification tasks.

Although the minimum data length of $I_{SG}$ and $I_{18}$ is $14$, we further access the performance of the classifiers on shorter replicates by removing 7 data points (i.e., ending the data one week before the outbreak timing) from these two COVID-19 datasets. Results presented in Supplementary Tables~\ref{18CovidResult_Shorter}-\ref{Singapore_Shorter} show that if the disease outbreak occurs seven days later, there is no significant difference between using the current data for prediction and additionally incorporating data from the next seven days. Thus, we can effectively predict the outbreak with a lead time of one week. It is worth noting that the “N” replicates are selected from the subset of SARS data with $R_e<1$, with random lengths, hence we do not conduct the same test on shorter replicates for this dataset. We should also acknowledge that asymptomatic or presymptomatic individuals exhibit scant influence on flu transmission~\cite{patrozou2009does}, in contrast, COVID-19 incidence data are significantly underestimated due to these “unnoticeable” infections. We expand the empirical data on individuals infected with COVID-19 five-fold and repeat the experiment. The results in Supplementary Tables~\ref{18CovidResult_Larger} and \ref{Singapore_Larger} show that underestimation of incidence data does not have a significant impact on classification performance, no matter whether the original performance is excellent or undesirable. 

\section{Discussion}\label{discussion}
Timely predictions of disease outbreaks can expedite the decision-making process and thus minimize associated losses. That said, implementing proactive and stringent actions when forecasting impending disease outbreaks; or promoting conservative and relatively mild policies when predicting non-outbreaks. Nevertheless, current efforts in early predictions of disease outbreaks and non-outbreaks, while urgent, are still far from perfect.

In our work, we introduce a novel framework for early forecasting the outbreaks (replicates labeled as “T”) and non-outbreaks (replicates labeled as “N”) using feature-based time series classification approaches. We train 32 classifiers following the framework, all achieving near-perfect classification performance. Further experiments show that these classifiers can handle time series of varying lengths and those far from the transition point (that is, outbreak timing). Notably, two classifiers, trained from 5EWSI of sliced time series and logistic regression model, exhibit an accuracy of $1$ over two empirical testing sets that are never used in the training process. These results may suggest that predicting impending transitions, or disease outbreaks in the context of epidemiology, can be achieved by a less sophisticated method. 

Early prediction of transitions is not new. Numerous novel EWIs, as well as their combinations, are proposed for transition prediction. Tredennick et al.~\cite{tredennick2022anticipating} detect CSD using EWIs in the context of measles transmission in Niger; Harris et al.~\cite{harris2020early} demonstrate that several statistical features exhibited specific characteristics before the outbreak of malaria in Kericho, Kenya; O’Brien and Clements~\cite{o2021early} access the reliability of EWSs through daily COVID-19 incidence sequences of 24 countries. Moreover, Bury et al.~\cite{bury2021deep} apply deep learning algorithms to classify different types of bifurcations using simulated data. However, disease-related applications of EWS have been conducted with outbreak cases, while cases where non-outbreaks occur are overlooked. In addition to achieving excellent performance in the detection of impending pandemics, we also aim to undermine the misjudgment of normality cases, i.e. reduce the type 2 error. In this study, we simulated time series data undergoing transcritical bifurcation (outbreak) and null bifurcation (non-outbreak) and achieved a classification accuracy exceeding $0.99$ in these simulations.

The limited length of time series, which is common in incidence data, is also a lingering problem for early prediction of outbreaks~\cite{o2021early, dablander2022overlapping}. We should further consider the trade-off between earliness and accuracy. Note that we can manipulate the length and location of the subsequences of the raw time series. With replicates farther from the outbreak timing, we observe consistently less performative results for classifiers except those trained from $I_W$ or $I_D$ of 5EWSI. This result is consistent with the CSD phenomenon that inner stability decreases and data fluctuation becomes more pronounced when approaching the transition point, resulting in the difference with those far from the transition point. Data will be more informative when closer to the transition points. For extremely short training data, particularly those with lengths less than $20$, classifiers exhibit lower AUC scores due to insufficient information learned from feature extraction. Still, they outperform the random chances. As the length increases, the AUC increases rapidly at first and then tends to stabilize. Similar to our work, Nanopoulos et al.~\cite{nanopoulos2001feature} apply four statistical features (the mean, standard deviation, skewness, and kurtosis) of the time series to conduct the TSC tasks, and claim that feature-based methods are more robust compared to valued-based ones.

Delecroix et al.~\cite{delecroix2023potential} retrieve studies on infectious disease outbreaks anticipation via EWIs, and find that more than half are conducted on simulated data only. It is premature to conclude the practicability purely based on simulations. In addition to applying synthetic data generated from stochastic differential equations (SDE) models, we assess the performance of classifiers on empirical datasets to predict whether outbreaks will occur. Although the classifiers are not trained on empirical datasets, some classifiers trained from synthetic data can still make correct classifications, as illustrated in Figure~\ref{FullAccuracy}.

It is noteworthy that the performance of the models for the two types of data seems to be complementary, even though the training sets are balanced. The other finding in performance is that 5EWSI, with lower dimensionality than 22SF, outperforms 22SF on real-world testing data, despite both sets of features in this paper achieving near-perfect classification accuracy and AUC scores of approximately 1 on synthetic data. This discrepancy in performance may be attributed to model overfitting using 22SF. Moreover, classifiers trained from the KNN model demonstrate outstanding performance over withheld testing sets from synthetic and empirical data labeled as “T” (COVID-19 data from Singapore $I_{SG}$). However, these classifiers consistently failed to classify “N” data (SARS data from Hong Kong $I_{HK}$). The KNN model classifies the new objective based on its “distance” from the training set, essentially storing instances of the training set without constructing a model. This may also elucidate why classifiers derived from KNN achieved good performance on withheld testing sets: for the KNN model, empirical “N” data is “closer” to simulated “T” data.

Generic EWS predictive ability is limited~\cite{hastings2010regime, dakos2015resilience, southall2021early}, but identifying mechanistic behavior~\cite{gsell2016evaluating} and incorporating specific information can enhance the prediction performance. Thomson et al.~\cite{thomson2001development} show the efficacy of early warning systems for malaria-outbreak anticipation using meteorological, epidemiological, and environmental factors. Climate- and weather-driven early warning systems can handle potential disease threats~\cite{mori2017early}. Our goal, however, is to make early predictions by assuming no prior knowledge of the infectious disease which is an unavoidable fact when a novel disease emerges. Classifiers derived from our proposed framework only require incidence data, the most straightforward data from surveillance.

There are several limitations in our work. The SIR model used in this work operates under the assumption of homogeneous mixing of the infected and susceptible populations. Such a prerequisite does not hold for large and complex communities (countries, for instance). During the simulation process, we assume a linear increase in the transmission rate over time. Future research will aim to address these limitations by exploring more realistic non-linear functions for the transmission rate. Additionally, although some of the trained classifiers can achieve an accuracy of more than 0.9 on empirical data, we only obtain the real-world data with either label “T” or label “N” as the testing set in our real case study. In other words, the testing set from the real-life records is highly biased, and we cannot compute the AUC scores of the classifiers. Furthermore, our work does not address two widespread concerns: (1) when the pandemic will emerge after the onset of the disease, and (2) how severe the potential pandemic could lead to without control policies. With an appropriate mathematical model, the former question can be addressed through effective reproduction number ($R_e$) estimation, while the latter can be theoretically answered by determining the final epidemic size. However, the dilemma lies in the fact that rigorous mathematical modeling itself requires a relatively longer time, and the assumptions of such the modeling process are counting on biological and epidemiological studies on the unknown disease. As a potential direction for future research, developing a more effective predictive approach on both outbreaks and non-outbreaks requires greater attention to address these gaps.

To conclude, our study reveals the inherent statistical differences in time series exhibiting outbreaks and non-outbreaks before their occurrence. Such discrepancies, in turn, enable the classification of these two scenarios proceeding the outbreak or non-outbreak happens. Thus, we introduce a framework for early predictions of disease outbreaks using the feature-based TSC method, and results demonstrate that all classifiers can accurately predict the outbreaks and non-outbreaks in a simulated context, with two achieving an accuracy of 1 on out-of-sample empirical data. Our work presents a less sophisticated yet effective approach to predicting disease outbreaks and non-outbreaks, addressing the realistic requirements of early disease emergence stages.

\section{Methods}
\subsection{Time series classification and predictive models}
\label{models}
For a univariate time series $I(t)$, feature-based methods perform a feature extraction from $I(t)$ before the classification procedure. The computed features serve to transform the time series into a structured format comprising $N$ numeric variables, allowing us to directly employ machine learning algorithms to classify the feature data, and subsequently accomplish TSC tasks. Here, we employ four predictive models, namely GBM, LRM, KNN, and SVM, for classification use. AUC scores~\cite{hajian2013receiver} of classifiers and classification accuracy on testing sets are recorded as performance measures.

In this work, the code and analysis are carried out using the \texttt{R}. Two models, GBM and KNN, are implemented using the \texttt{caret} package~\cite{kuhn2008building}. The SVM model is applied using the \texttt{e1071} package~\cite{dimitriadou2008misc}. The LRM model, which is the logistic regression model in our work, is computed by the \texttt{LRM} function contained in the \texttt{stats} package of \texttt{R}~\cite{team2010r}. The AUC of each classifier is calculated using the \texttt{pROC} package~\cite{robin2011proc}.

\subsection{Synthetic data used for training and testing}
We first elaborate on how the synthetic data are simulated from dynamic models. Then we explain how the simulated time series are pre-processed for classification use. Note that synthetic data are used for both the training and testing processes in this paper.

\subsubsection{Synthetic time series generation}
\label{Synthetic Time Series Generation}
We start the investigation on the classic susceptible-infectious-recovered (SIR) model~\cite{kermack1927contribution}:
\begin{equation}
\begin{cases}
\frac{dS(t)}{dt} = \Lambda -\beta(t) S(t)I(t)-\mu S(t), \\
\frac{dI(t)}{dt} = \beta(t) S(t)I(t)-\alpha I(t) -\mu I(t), \\
\frac{dR(t)}{dt} = \alpha I(t) -\mu R(t),
\end{cases}
\label{SIR}
\end{equation}
where $S$, $I$, and $R$ are susceptible, infected, and recovered individuals at time $t$, respectively. $\Lambda$ is the recruitment rate of susceptible population, $\beta(t)$ is the transmission rate at time $t$, $\mu$ is the death rate, and $\alpha$ is the recovery rate. 

Admittedly, the SIR model is often regarded as simplistic in capturing the dynamics of infectious diseases. More complex and context-specific models are better suited for simulating the intricate transmission mechanisms in real-world scenarios. But it is also the simplicity of the SIR model that provides the “greatest common factor” to accommodate the novel disease to some extent. Furthermore, the dynamics of infectious diseases can be easily influenced by random internal and/or external disturbances. To augment the SIR model, we followed Chakraborty et al. \cite{chakraborty2024early} and introduced three distinct stochastic differential equations (SDEs) incorporating white noise, multiplicative environmental noise, and demographic noise to generate synthetic incidence time series. The rationale behind selecting these specific noise types, as well as the parametrization, are justified in Chakraborty et al.\cite{chakraborty2024early}. 

Mathematically, the basic reproductive number $R_0$, a dimensionless value representing the expected count of secondary infections caused by a single infectious individual in a completely susceptible population, serves as an important metric for the stability of equilibrium points in the SIR model~\cite{fraser2009pandemic}. When $R_0<1$, the disease-free equilibrium $E_1=(\Lambda, 0, 0)$ is stable, and the system remains in a non-outbreak state. $R_0>1$ indicates the potential for the disease to persist within the population, and the endemic equilibrium $E_2=(\frac{\mu+\alpha}{\beta}, \frac{\Lambda}{\mu+\alpha}-\frac{\mu}{\beta}, \frac{\alpha\Lambda}{\mu(\mu+\alpha)}-\frac{\alpha}{\beta})$ becomes stable. At the critical threshold of $R_0=1$, a transcritical bifurcation occurs, wherein the disease-free equilibrium and endemic equilibrium intersect and exchange their stability properties. Following the methodology proposed by Chakraborty et al. \cite{chakraborty2024early}, all parameters except the transmission rate $\beta$ are to be constants. Then the value of $R_0$ becomes solely dependent on $\beta(t)$, expressed as $R_0(t) = K\beta(t)$, with $K=\frac{\Lambda}{\mu (\alpha + \mu)}$. We further assume a linear variation in the transmission rate $\beta$ with respect to time $t$: $\beta(t) = \beta_0 + \beta_1 t$, where $\beta_0$ and $\beta_1$ follow triangular distributions. If the slope $\beta_1$ is sufficiently small, not exceeding the critical value of reproduction number within the time interval $t \in [1, 1500]$, the simulation results in null bifurcation. Otherwise, we obtain a simulation with a random transition time within the time interval $t \in [1, 1500]$.

\subsubsection{Synthetic data input}
\label{DataGeneration}
\textbf{WhiteN (EnvN, DemN, or MixedN) (\# replicates = 14,400; Data simulated from SIR model with noise)}
Following the simulation process outlined in section ~\ref{Synthetic Time Series Generation}, we generate three datasets from three corresponding SDEs, denoted as WhiteN (SIR model with white noise), EnvN (SIR model with multiplicative environmental noise), and DemN (SIR model with demographic noise). For each SDE model, 7,200 replicates $I[1:1500]$ (univariate time series) exhibiting transcritical bifurcation are generated. Note that replicates with transition time points $T$ smaller than 400 are excluded to avoid inconsistencies in the following data-slicing process. An additional 7,200 replicates $I[1:1500]$ where no bifurcation event occurs are also generated from each SDE model. Furthermore, from each category in the three datasets, we randomly select 2,400 replicates, forming a new dataset named MixedN.

Since we aim to predict the outbreak before its occurrence, it is imperative to exclude the data after the transition point $T$ for full-time series $I[1:1500]$ exhibiting transcritical bifurcation when forming the training sets and testing sets. Therefore, we extract $400$ data points before the transition point and label it as “T”, as illustrated in Figure~\ref{SummarySynthetic} (b). To maintain consistency, $400$ consecutive data points are randomly chosen for each simulation with null bifurcation. A summary of the synthetic data used in this study is described in Figure~\ref{SummarySynthetic}.

\subsection{Empirical data used for testing}
\label{Empirical Data Used for Testing}
To assess the applicability of the framework in real scenarios, we apply empirical time series data of two infectious diseases: the novel coronavirus disease (COVID-19) and severe acute respiratory syndrome (SARS). COVID-19 has emerged as an excellent subject for novel disease studies due to the abundance of accessible data, both temporally and spatially, and its prolonged duration. COVID-19 exemplifies novel infectious diseases causing significant outbreaks (transcritical bifurcation). The SARS outbreak was initially identified in November 2002~\cite{zhong2003epidemiology}. Since 2004, there have been no known cases of SARS reported worldwide, providing an illustrative example of non-outbreak (null bifurcation) in our study.

Similar to synthetic data, we apply the effective reproduction number ($R_e$) as the criterion to determine whether empirical time series exhibiting outbreaks (transcritical bifurcation) and the timing of outbreak occurrence. Given a full-time series $I$ exhibiting outbreak(s), $I[k_0:k]$ is labeled as “T” if and only if $R_e[t]<1$ for any $k_0\leq t\leq k$, $R_e[k_0-1]\geq1$, and $R_e[k+1]\geq1$. For $I$ exhibiting the non-outbreak, any subsets will be labeled as “N”. 

\textbf{$I_{SG}$ (\# replicates = 19; COVID-19 daily infection data from Singapore)} The fundamental assumption of the classic SIR model is the homogeneity of population distribution (i.e. each individual in the population has an equal contact probability). However, this assumption may not hold for data spanning a geographically extensive area, such as national-level data, despite their potential for greater accuracy and availability. Nonetheless, the COVID-19 data from Singapore presents an ideal subject for study, where the population demographics align well with the assumptions of the SIR model. Additionally, the government responded promptly to the pandemic, implementing intensive tracking via mobile technology (Government Technology Agency of Singapore~\cite{Government}), enhancing data accuracy. To maintain the consistency of the SIR model and the empirical testing set, we focus on the COVID-19 infection data from Singapore, spanning from January 26, 2020, to February 18, 2024. The data is collected weekly and then smoothed by evenly distributing the infection number over seven days of the corresponding week to obtain daily data. $R_e$ is estimated via \texttt{R} package \texttt{EpiEstim}~\cite{cori2020package} with a mean value of $6.3$ and a standard deviation of $4.2$ for the serial interval~\cite{nash2023estimating}. Replicates with lengths shorter than $14$ days are excluded, resulting in a testing set comprising 19 replicates labeled as “T”.

\textbf{$I_{HK}$ (\# replicates = 1200; SARS 2003 incidence data from Hong Kong)} We utilize SARS data from Hong Kong, spanning from March 17 to July 11, 2003~\cite{SARS2003}. The original data comprises Cumulative Cases, Death Cases, and Recovered Cases. To maintain consistency with the SIR model, we computed the prevalence (i.e., the “I” term in the SIR model) by subtracting “Death Cases” and “Recovered” from Cumulative Cases. The recorded data displays missing values, with one missing data point every Sunday for the initial 11 weeks and two missing data points every Saturday and Sunday in the last 5 weeks (see Figure~\ref{HK_Data_Re} (a)). To impute the missing data, we applied \texttt{R} package \texttt{imputeTS} using \texttt{na\_interpolation} and obtained the completed incidence time series. The mean and standard deviation of the serial interval are 8.4 and 3.8, respectively~\cite{lipsitch2003transmission}. We found that the mean value of estimated \textbf{$R_e$} falls below 1 during the estimation interval from April 18 to April 24, 2003, and never exceeds $1$ afterward. Then we designate the sequence after April 24 2003 as the null bifurcation scenario and randomly select 1200 subsets, each with random lengths larger than $7$, as the “N” replicates in the testing set.

\subsection{Statistical features and early warning signal indicators}
Classifying raw time series, especially those observed at higher frequencies, usually involves working with high-dimensional data, which can be computationally intensive and less interpretable~\cite{liao2005clustering}. To address this concern, people seek a proper representation of the original time series, typically encompassing statistical attributes, in a numerical format. In this paper, we employ two distinct sets of features, 22SF and 5EWSI, to represent the sequence for classification purposes. 

Numerous alternative features for time series are available, and among them, 22SF computed through \texttt{R} package \texttt{Rcatch22} stands out as a promising choice, as it exhibits ideal performance in TSC and minimal redundancy~\cite{lubba2019catch22}. The 22SF utilized in this study are detailed in Supplementary Table~\ref{22F}, and more information is available in Lubba et al.~\cite{lubba2019catch22}. Note that the 22SF of the simulated time series do not follow normal distribution, therefore, we conducted the Mann-Whitney U test on these features computed from “T” and “N” sequences across four types of noise data. The results are depicted in Supplementary Figures~\ref{boxW}-\ref{boxM}. As can be easily observed, the features computed from data with two labels are significantly different, with $p\ll0.001$. These findings suggest that despite the visually indistinguishable nature of $I_{Data}^T$ and $I_{Data}^N$, these two types of sequences can be theoretically separated by examining these statistical features.

Alternatively, we consider 5 early warning signal indicators (5EWSI) from time series to serve as the features, with their formulas provided in Supplementary Table~\ref{5S}. Notably, there is no overlapping between 22SF and 5EWSI. We also implemented the Mann-Whitney U test, and results demonstrate that there is a difference in these five features computed from time series labeled as “T” and “N” (Supplementary Figures~\ref{SboxW}-\ref{SboxM}).

\subsection{Experiment settings for earliness and accuracy}
Maximizing classifier accuracy using the complete time series is typically the primary goal of TSC problems~\cite{gupta2020approaches}. However, in time-sensitive applications, such as infectious disease monitoring and pandemic forecasting, the aim shifts to optimizing the balance between two contradictory (or contradicting or opposing) objectives: accuracy and earliness in the classification task. Such transition implies that the classifier should ideally handle inadequate (shorter time intervals) and early (those far from the impending transition) sequences while maintaining a certain level of classification accuracy to provide reliable references for policymakers. Consequently, critical questions arise in two-fold: How early can predictions be reliably provided? What is the sufficient length of testing time series required to offer a reliable prediction? To address these questions, we conducted two experiments, as illustrated in Figure~\ref{RollingAUC} (a) and Figure~\ref{ExapndingAUC} (a). 

Firstly, we employ the Rolling window approach on both $I_{Data}^T$ and $I_{Data}^N$. We maintain a fixed length of $100$ for input time series $L$ and “roll” the $L$ from right to left, resulting in an increasing gap $D$ between the last time point of the window and the transition point $T$. Starting with an initial gap of $0$, we increment the gap by $5$ units for each iteration $k$, yielding $61$ experiments. After determining the input time series $L$ for each round, we follow the same feature-computation and data partition procedure outlined in section ~\ref{Synthetic Time Series Generation} and train classifiers through four models. We record the AUC and classification accuracy on testing sets for each classifier at every iteration.

To further investigate predictive performance using varying lengths of the time series data, we conduct training and testing processes using an Expanding window approach. Initiating with a 5-time-point input time series $L$, we expand the $L$ by adding five more data points to the left end of the time series for each iteration, while maintaining the length of gap $D$ as $30$. Consequently, the length of input time series $L$ ranges from $5$ to $370$, resulting in $74$ reiterations for each time series dataset and each model. We again follow the feature extraction, and training-testing sets partition approaches, and perform the classification. We repeat these experiments across all eight datasets and four models, recording the corresponding AUC scores and accuracy values for assessments.

\section*{Data availability}
Synthetic data are available at \url{https://doi.org/10.5281/zenodo.10967222}~\cite{Gao_2024_10967222}. COVID-19 incidence data from Singapore is from Our World in Data~\cite{owidcoronavirus}.
The SARS data for Hong Kong were obtained from \url{https://www.kaggle.com/datasets/imdevskp/sars-outbreak-2003-complete-dataset?resource=download}~\cite{SARS2003}.

\section*{Code availability}
Code to reproduce the data-processing, experiments, analysis, and figures is available at \url{https://doi.org/10.5281/zenodo.10967222}~\cite{Gao_2024_10967222}.

\section*{Acknowledgments}
This project (S.G. and A.K.C) was primarily supported by One Health Modelling Network for Emerging Infections (OMNI), Amii (Alberta Machine Intelligence Institute) matching fund, and the Department of Mathematical and Statistical Sciences (IUSEP funding) at the University of Alberta. H.W. was partially supported by the Natural Sciences and Engineering Research Council of Canada (Individual Discovery Grant RGPIN-2020-03911 and Discovery Accelerator Supplement Award RGPAS-2020-00090) and the CRC program (Tier 1 Canada Research Chair in Mathematical Biosciences). M.A.L gratefully acknowledges support from an NSERC Discovery Grant and the Gilbert and Betty Kennedy Chair in Mathematical Biology. R.G. was funded by NSERC and CIFAR. We thank Pouria Ramazi, Tianyu Guan, Reza Miry, Ilhem Bouderbala, and Russell Milne for their valuable feedback.

\section*{Author contributions}
S.G., H.W., and M.A.L. conceptualized this study. R.G. contributed to refining the ideas. A.K.C. conducted the simulation. H.W. and M.A.L. provided project supervision. S.G. wrote the original draft. All authors revised and commented on the
manuscript.

\section*{Conflict of interest}
The authors declare that they have no conflict of interest.

\newpage
\section*{\large Supplementary Materials}
\setcounter{section}{0}
\setcounter{table}{0}  
\setcounter{figure}{0} 

\renewcommand{\thetable}{S\arabic{table}}
\renewcommand{\thefigure}{S\arabic{figure}}
In this document, we provide our supporting information for the manuscript entitled “Early detection of disease outbreaks and non-outbreaks using incidence data”. We start by presenting three stochastic differential equations (SDEs) employed for data simulation. Then we elaborate on two feature extraction libraries (22 statistical features and 5 early warning signal indicators). We further present supplemental results.

\section{Data generation}
Within the framework of the SIR model, we incorporated three distinct sources of stochastic variation, namely white noise, multiplicative environmental noise, and demographic noise, to generate the datasets. These models are shown below:

SIR model with White Noise
    \begin{equation}
    \begin{aligned}
     dS&=\Lambda d t-\beta(t) S I d t-\mu S d t+\sigma_1 d W_1 \\
     dI&=\beta(t) S I d t-\alpha I d t-\mu I d t+\sigma_2 d W_2
     \end{aligned}
     \label{SDE}
    \end{equation}
    where $\sigma_1$ and $\sigma_2$ are noise intensity and $W_i(t),$ $i = 1, 2,$ are the Wiener process.

SIR model with Multiplicative Environmental Noise
    \begin{equation}
        \begin{aligned}
    d S & =\Lambda d t-\beta S I d t-\mu S d t+\sigma_1 S d W_1 \\
    d I & =\beta S I d t-\alpha I d t-\mu I d t+\sigma_2 I d W_2
    \end{aligned}
    \end{equation}
    where $\sigma_1$ and $\sigma_2$ are noise intensity and $W_i(t),$ $i = 1, 2,$ are the Wiener process.

SIR model with Demographic Noise
    \begin{equation}
      \begin{aligned}
 \displaystyle{dS(t)} &= \displaystyle{\Lambda dt -\beta(t) S(t)I(t) dt-\mu S(t) dt+ \frac{a(t)+d(t)}{e(t)} dW_1(t)+\frac {b(t)}{e(t)} dW_2(t)}, \\
\displaystyle{dI(t)} &= \displaystyle{\beta(t) S(t)I(t) dt-\alpha I(t) dt - \mu I(t) dt + \frac{b(t)}{e(t)} dW_1(t) + \frac{c(t)+d(t)}{e(t)} dW_2(t)}, \\ 
      \end{aligned}
    \end{equation}
where
$a(t) = \displaystyle{\Lambda +\beta(t) S(t)I(t) +\mu S(t)}$,  
$b(t) = -\beta(t) S(t) I(t)$, 
$c(t) = \beta(t) S(t)I(t) + \alpha I(t) + \mu I(t)$, 
$d(t) = \sqrt{a(t) c(t) - b^2(t)}$, 
$e(t) =\sqrt{a(t)+c(t)+2d(t)}$, 
and $W_i(t),$ $i = 1, 2,$ are the Wiener process.

We simulate 14400 time series from each of these three SDEs, with half exhibiting transcritical bifurcation (outbreak) and the rest showing null bifurcation (normality). Additionally, we compile one dataset, MixedN, by randomly choosing 2400 time series for each scenario from the other three datasets. The summary is provided in Table \ref{Train_Test_Unbalanced}:
\begin{table}[H]
\centering
\begin{tabular}{cccccc}
\toprule
&  & WhiteN & EnvN & DemN & MixedN \\
\hline
\multirow{2}{*}{\textbf{Training}}  & Transcritical & 6000 & 6000 & 6000 & 6000 \\
& Null & 6000 & 6000 & 6000 & 6000 \\
\hline
\multirow{2}{*}{\textbf{Testing}}  & Transcritical & 1200 & 1200 & 1200 & 1200 \\
 & Null & 1200 & 1200 & 1200 & 1200 \\
\bottomrule
\end{tabular}
\caption{Summary of Training set and withheld Testing set.}
\label{Train_Test_Unbalanced}
\end{table}

\newpage
\section{22 statistical features and Mann-Whitney U Test results over synthetic data}
\begin{figure}[H]
    \centering
    \includegraphics[width=\textwidth]{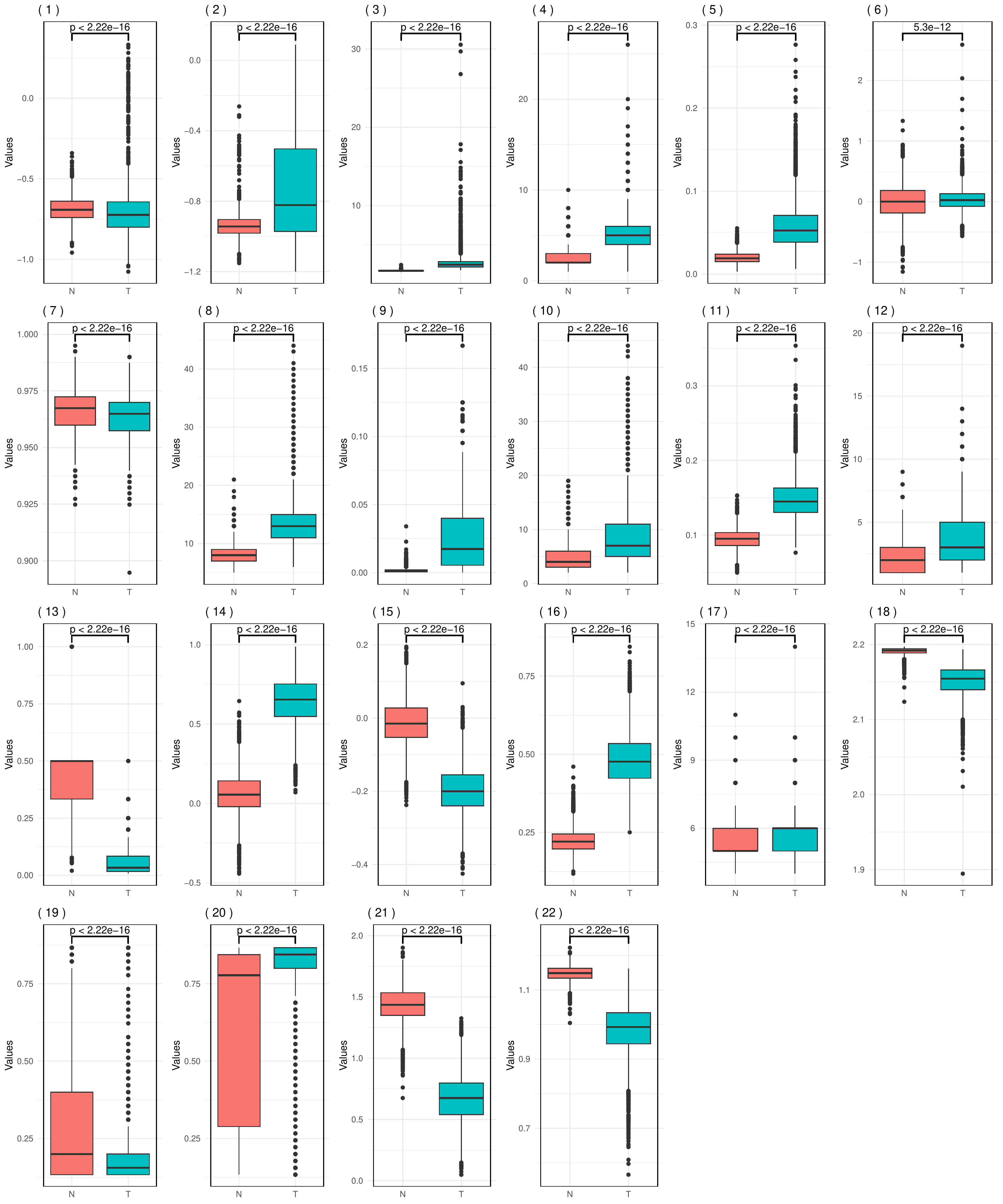}
    \caption{Features computed for the two types of generated data (white noise). N indicates null bifurcation data and T indicates transcritical bifurcation data. There are 22 time series features, see Table \ref{22F} for details. P-Values for 22 Features using Mann-Whitney U Test.}
    \label{boxW}
\end{figure}

\begin{figure}[H]
    \centering
    \includegraphics[width=\textwidth]{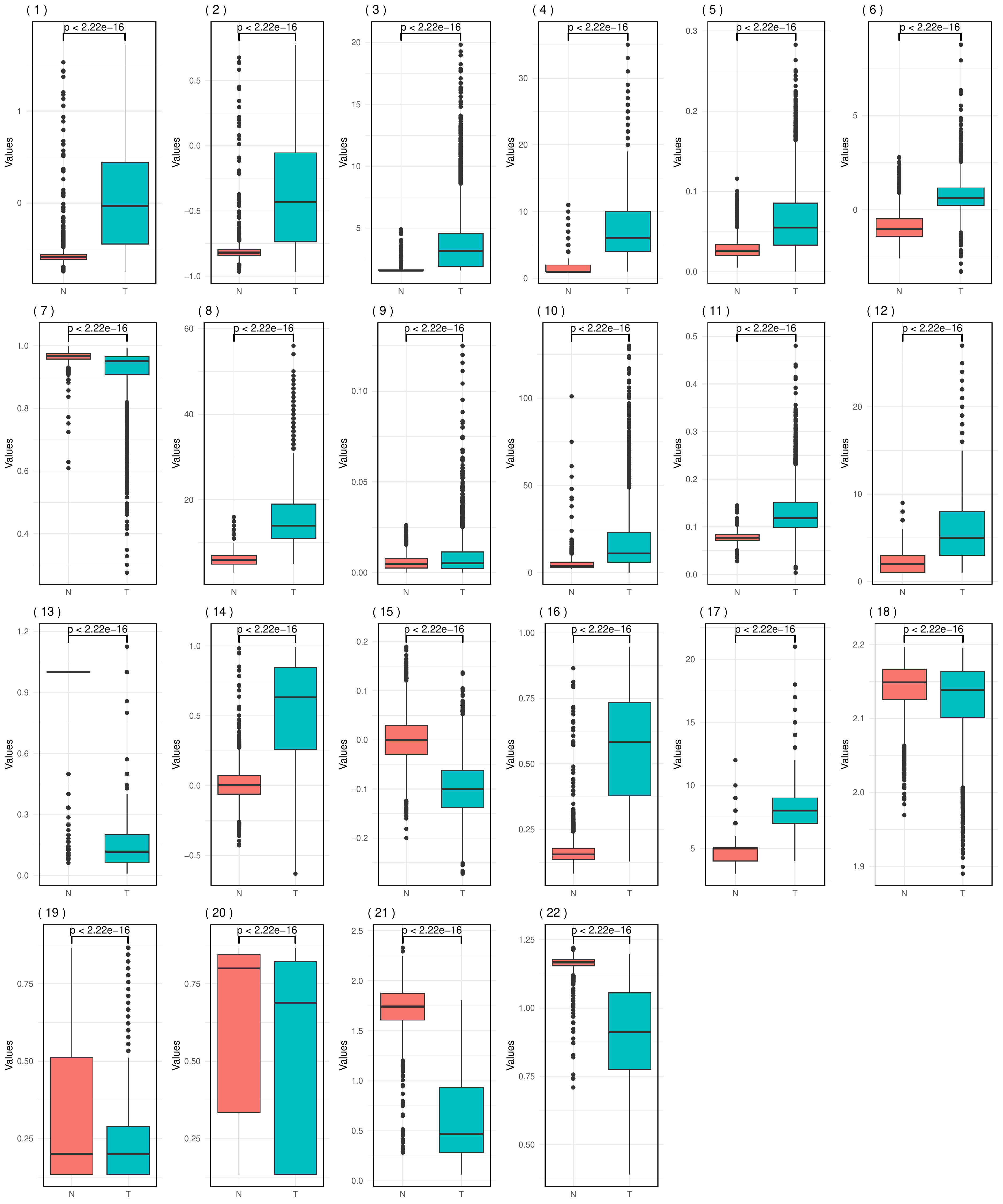}
    \caption{Features computed for the two types of generated data (environmental noise). N indicates null bifurcation data and T indicates transcritical bifurcation data. There are 22 time series features, see Table \ref{22F} for details. P-Values for 22 Features using Mann-Whitney U Test.}
    \label{boxE}
\end{figure}

\begin{figure}[H]
    \centering
    \includegraphics[width=\textwidth]{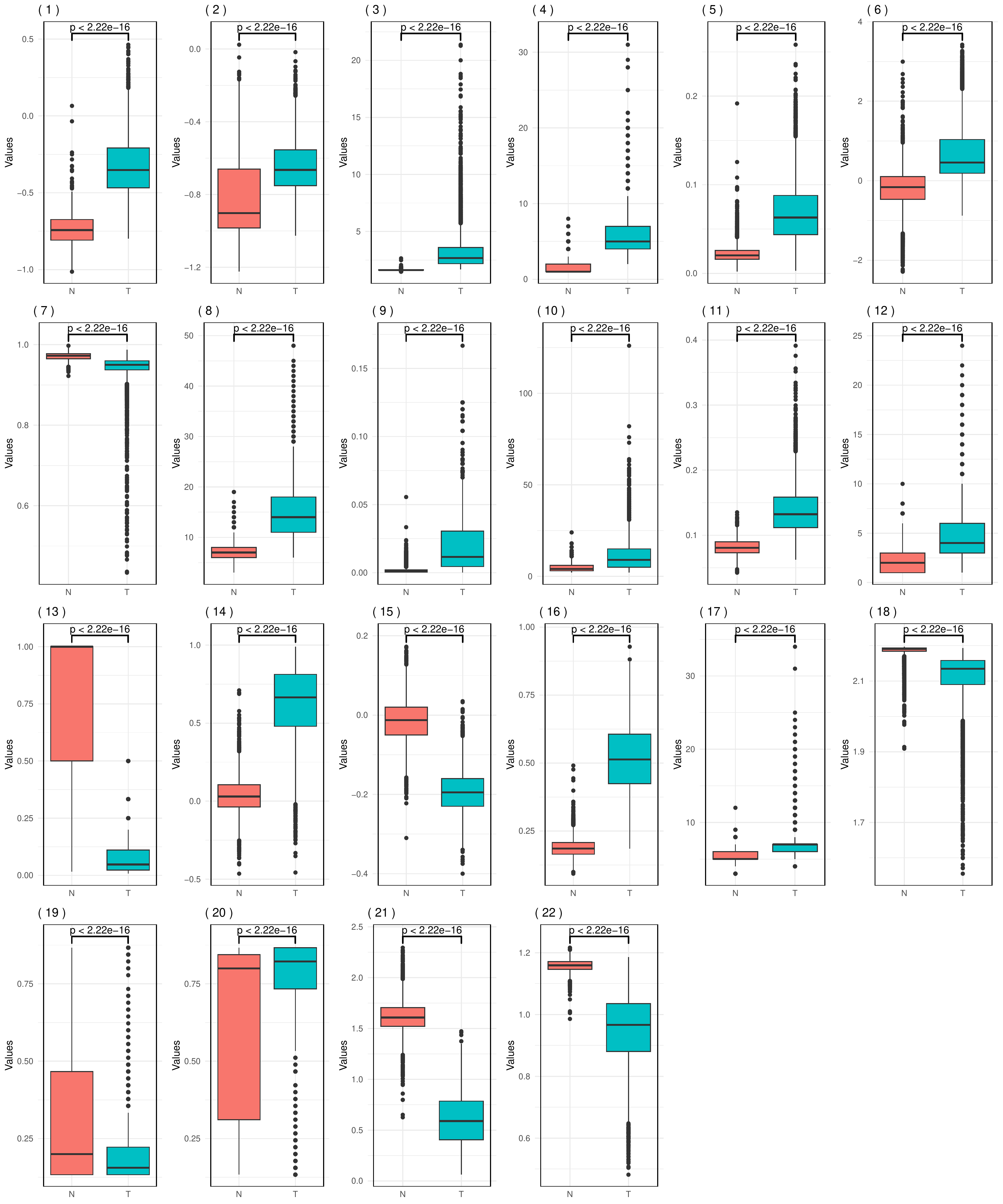}
    \caption{Features computed for the two types of generated data (demographic noise). N indicates null bifurcation data and T indicates transcritical bifurcation data. There are 22 time series features, see Table \ref{22F} for details. P-Values for 22 Features using Mann-Whitney U Test.}
    \label{boxD}
\end{figure}

\begin{figure}[H]
    \centering
    \includegraphics[width=\textwidth]{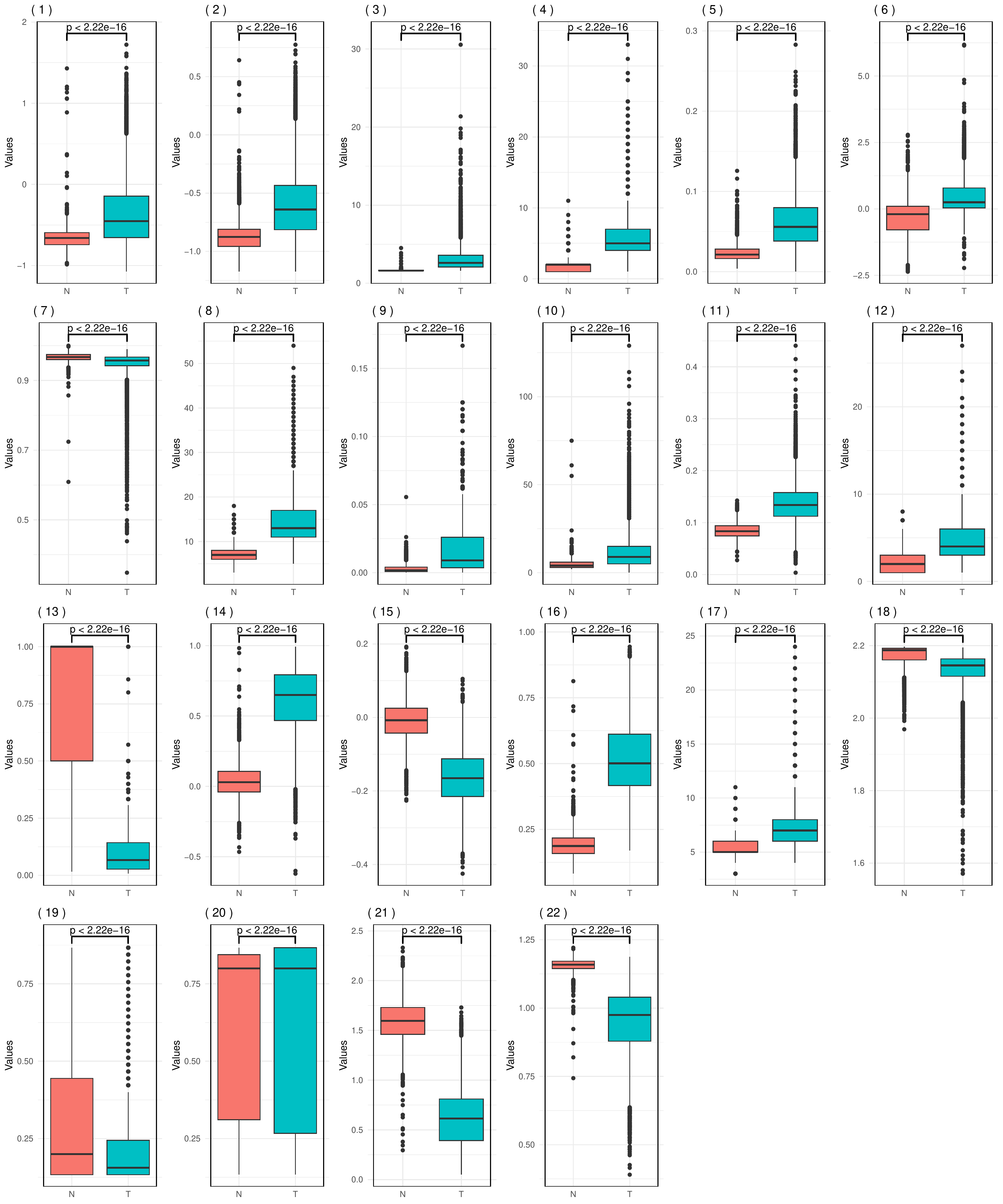}
    \caption{Features computed for the two types of generated data (mixed noise). N indicates null bifurcation data and T indicates transcritical bifurcation data. There are 22 time series features, see Table \ref{22F} for details. P-Values for 22 Features using Mann-Whitney U Test.}
    \label{boxM}
\end{figure}

In this section, we present detailed descriptions of 22 statistical features \cite{lubba2019catch22} in Table \ref{22F}. Then the box plots of each feature computed from “T” and “N” samples are provided in Fig \ref{boxW}-\ref{boxM}. Since the feature distributions are skewed in most cases, we perform the Mann-Whitney U Test, a non-parametric that does not assume specific data distribution, to compare the differences between each feature of "T" and "N" samples.
{\renewcommand{\arraystretch}{0.9} 
\begin{table}[H]
\centering
\begin{tabular}{p{0.5cm} p{6.7cm} p{8.5cm}}
\toprule
\textbf{No.} & \textbf{Feature} & \textbf{Description} \\
\midrule
1 & DN\_HistogramMode\_5 & Mode of z-scored distribution (5-bin histogram) \\
2 & DN\_HistogramMode\_10 & Mode of z-scored distribution (10-bin histogram) \\
3 & CO\_f1ecac & First $1/e$ crossing of autocorrelation function \\
4 & CO\_FirstMin\_ac & First minimum of autocorrelation function \\
5 & CO\_HistogramAMI\_even\_2\_5 & Automutual information, $m = 2$, $\tau = 5$  \\
6 & CO\_trev\_1\_num & Time-reversibility statistic, $\left\langle\left(x_{t+1}-x_t\right)^3\right\rangle_t$ \\
7 & MD\_hrv\_classic\_pnn40 & Proportion of successive differences exceeding $0.04\sigma$ \cite{mietus2002pnnx}\\
8 & SB\_BinaryStats\_mean\_longstretch1 & Longest period of consecutive values above the mean \\
9 & SB\_TransitionMatrix\_3ac\_sumdiagcov & Trace of covariance of transition matrix between symbols in 3-letter alphabet \\
10 & PD\_PeriodicityWang\_th0\_01 & Periodicity measure of \cite{wang2007structure} \\
11 & CO\_Embed2\_Dist\_tau\_d\_expfit\_meandiff & Exponential fit to successive distances in 2-d embedding space \\
12 & IN\_AutoMutualInfoStats\_40\_gaussian\_fmmi & First minimum of the automutual information function \\
13 & FC\_LocalSimple\_mean1\_tauresrat & Change in correlation length after iterative differencing \\
14 & DN\_OutlierInclude\_p\_001\_mdrmd & Time intervals between successive extreme events above the mean \\
15 & DN\_OutlierInclude\_n\_001\_mdrmd & Time intervals between successive extreme events below the mean \\
16 & SP\_Summaries\_welch\_rect\_area\_5\_1 & Total power in lowest fifth of frequencies in the Fourier power spectrum \\
17 & SB\_BinaryStats\_diff\_longstretch0 & Longest period of successive incremental decreases \\
18 & SB\_MotifThree\_quantile\_hh & Shannon entropy of two successive letters in equiprobable 3-letter symbolization \\
19 & SC\_FluctAnal\_2\_rsrangefit\_50\_1\_logi\_prop\_r1 & Proportion of slower timescale fluctuations that scale with DFA ($50\%$ sampling) \\
20 & SC\_FluctAnal\_2\_dfa\_50\_1\_2\_logi\_prop\_r1 & Proportion of slower timescale fluctuations that scale with linearly rescaled range fits \\
21 & SP\_Summaries\_welch\_rect\_centroid & Centroid of the Fourier power spectrum \\
22 & FC\_LocalSimple\_mean3\_stderr & Mean error from a rolling 3-sample mean forecasting \\
\bottomrule
\end{tabular}
\caption{The catch22 feature set spans a diverse range of time series characteristics representative of the diversity of interdisciplinary methods for time-series analysis \cite{lubba2019catch22}.}
\label{22F}
\end{table}
}

\newpage
\section{5 early warning signal indicators and Mann-Whitney U Test results over synthetic data}
\begin{figure}[H]
    \centering
    \includegraphics[width=0.9\textwidth]{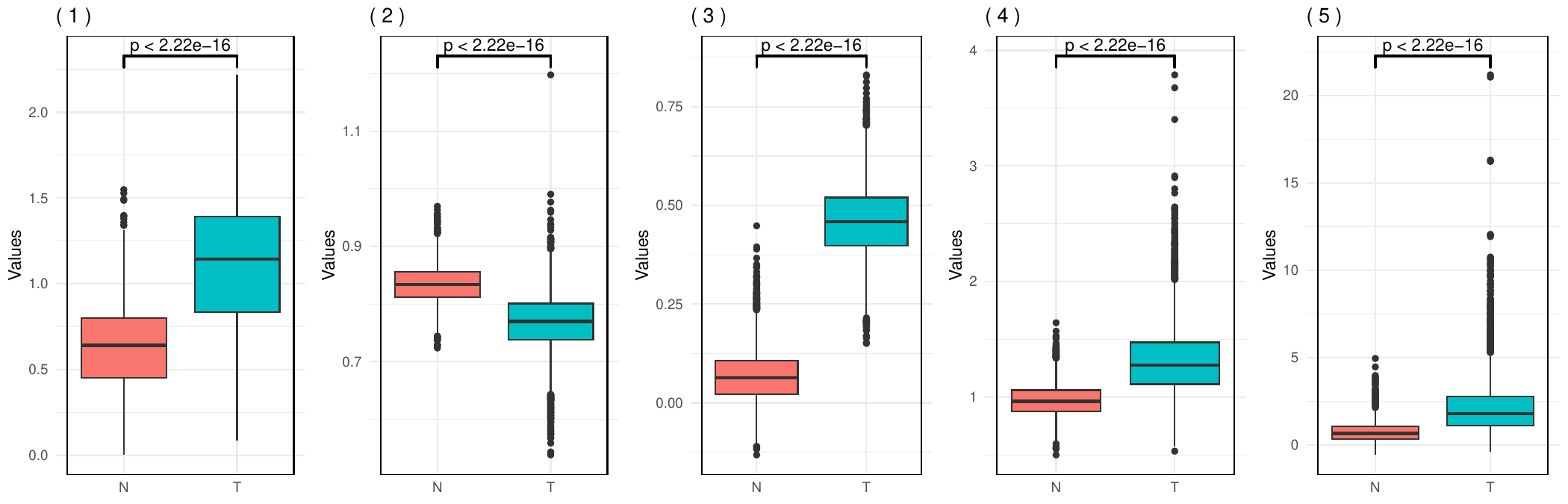}
    \caption{EWS indicators computed for the two types of generated data (white noise). N indicates null bifurcation data and T indicates transcritical bifurcation data. There are 5 EWS indicators, see Table \ref{5S} for details. P-Values for 5EWSI using Mann-Whitney U Test.}
    \label{SboxW}
\end{figure}

\begin{figure}[H]
    \centering
    \includegraphics[width=0.9\textwidth]{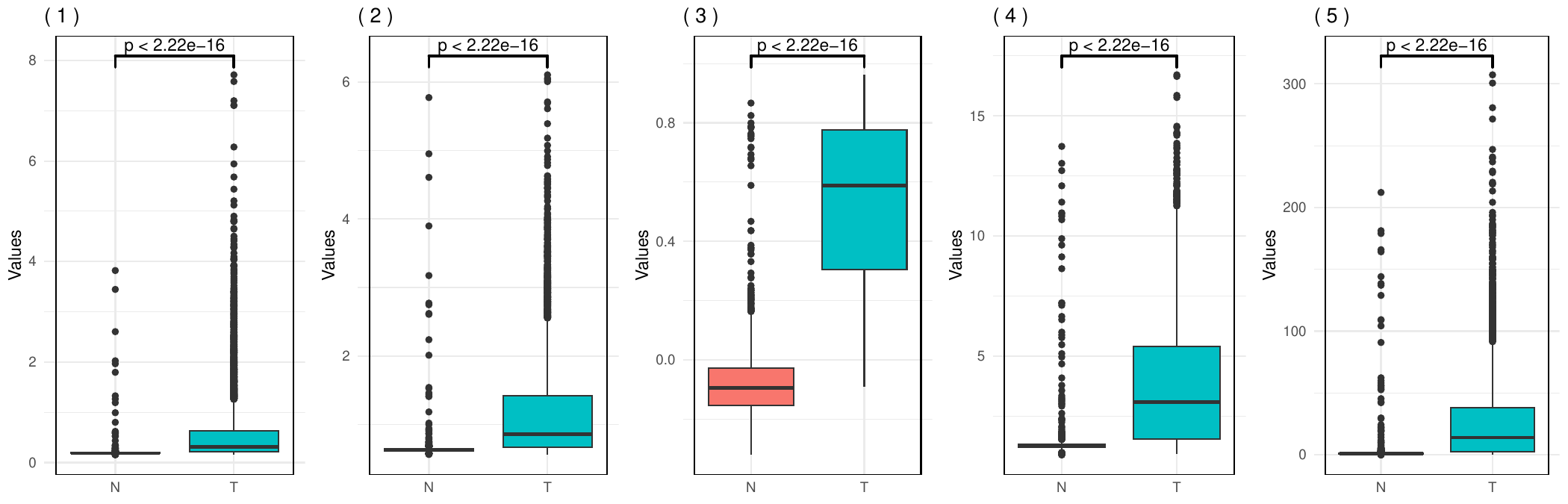}
    \caption{EWS indicators computed for the two types of generated data (environmental noise). N indicates null bifurcation data and T indicates transcritical bifurcation data. There are 5 EWS indicators, see Table \ref{5S} for details. P-Values for 5EWSI using Mann-Whitney U Test.}
    \label{SboxE}
\end{figure}

\begin{figure}[H]
    \centering
    \includegraphics[width=0.9\textwidth]{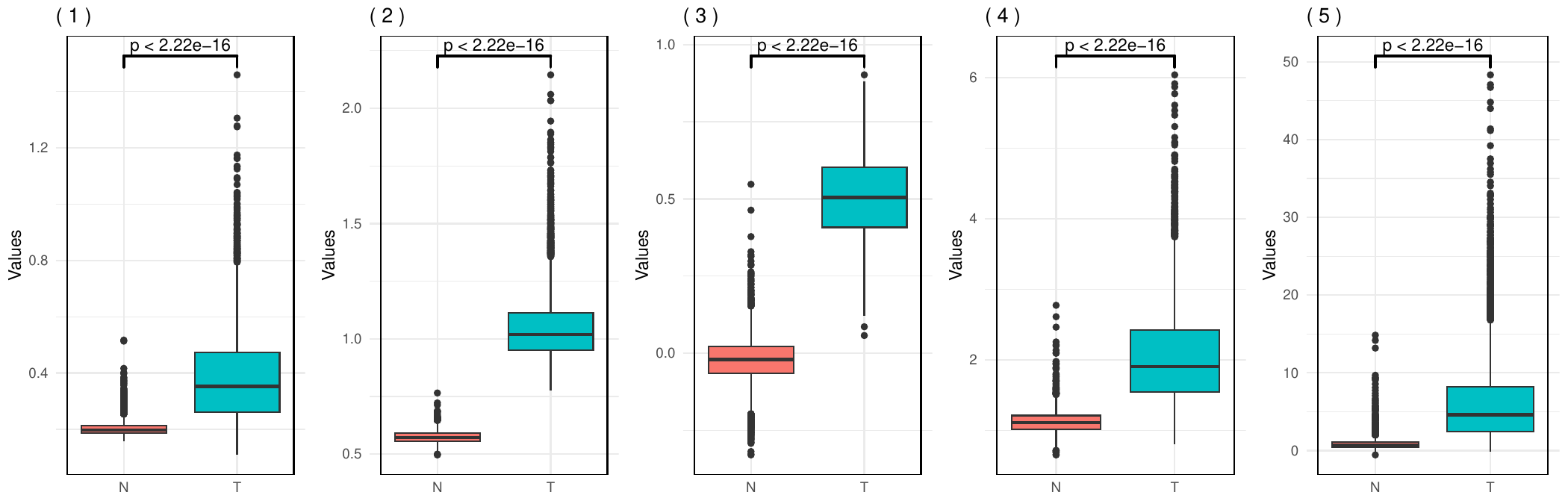}
    \caption{EWS indicators computed for the two types of generated data (demographic noise). N indicates null bifurcation data and T indicates transcritical bifurcation data. There are 5 EWS indicators, see Table \ref{5S} for details. P-Values for 5EWSI using Mann-Whitney U Test.}
    \label{SboxD}
\end{figure}

\begin{figure}[H]
    \centering
    \includegraphics[width=0.9\textwidth]{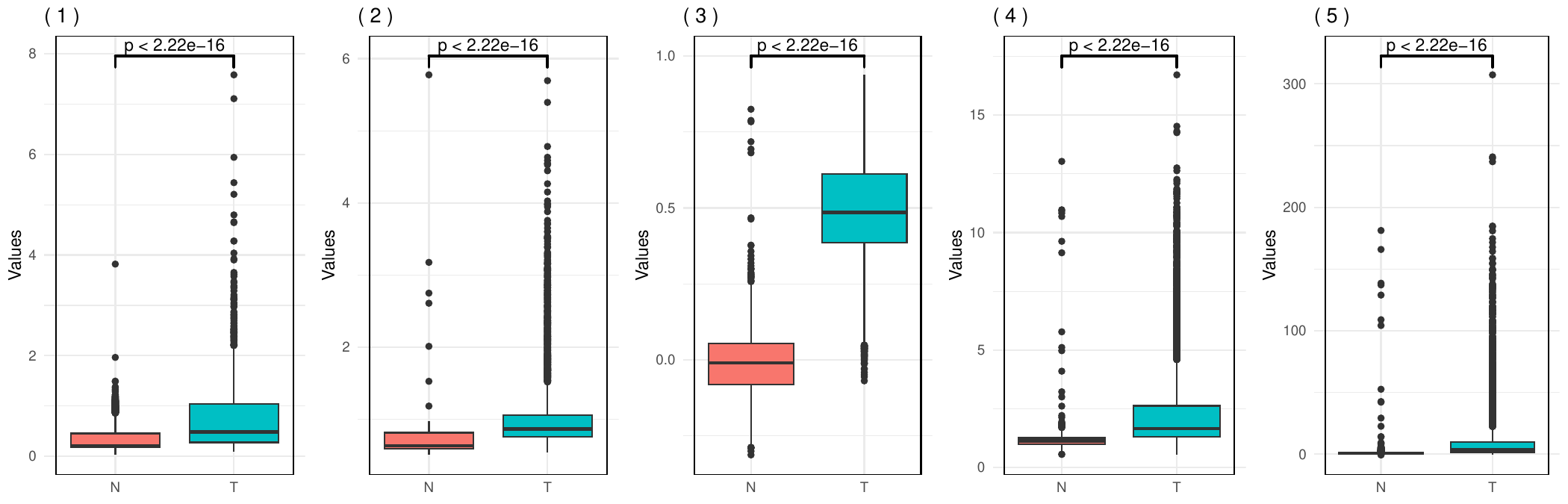}
    \caption{EWS indicators computed for the two types of generated data (mixed noise). N indicates null bifurcation data and T indicates transcritical bifurcation data. There are 5 EWS indicators, see Table \ref{5S} for details. P-Values for 5EWSI using Mann-Whitney U Test.}
    \label{SboxM}
\end{figure}

Here, we present detailed descriptions of 5 early warning signal indicators in Table \ref{5S}. Then the box plots of each feature computed from “T” and “N” samples are provided in Fig \ref{SboxW}-\ref{SboxM}. Similar to 22SF, we perform the Mann-Whitney U Test on each feature of "T" and "N" samples, as the features do not follow a normal distribution.
\begin{table}[H]
\centering
\small
\begin{tabular}{cccc}
\toprule
Indicator & Formula & Trend & Reference \\
\midrule
Standard Deviation (SD) & $\sigma(X) = \sqrt{\frac{\sum_{i=1}^N (x_i - \bar{x})^2}{N-1}}$ & Increase & \citep{carpenter2006rising} \\
\hspace{1em}\\
Coefficient of Variation (CV) & $\text{CV} = \frac{\sigma}{\bar{x}} \times 100\%$ & Increase & \citep{carpenter2006rising} \\
\hspace{1em}\\
Autocorrelation at Lag1 (AR1) & $AR^{(1)}(X) = \frac{1}{N-1} \sum_{i=1}^N \left(\frac{x_i - \bar{x}}{\sigma_x}\right)\left(\frac{y_i - \bar{y}}{\sigma_y}\right)$ & Increase & \citep{held2004detection} \\
\hspace{1em}\\
Skewness & $S(X) = \frac{\sum_{i=1}^N (x_i - \bar{x})^3}{N\sigma(X)^3}$ & Increase or Decrease & \citep{guttal2008changing} \\
\hspace{1em}\\
Kurtosis & $\text{Kurtosis} = \frac{\frac{1}{N}\sum_{i=1}^N (x_i - \bar{x})^4}{\left(\frac{1}{N}\sum_{i=1}^N (x_i - \bar{x})^2\right)^2}$ & Increase & \citep{biggs2009turning} \\
\bottomrule
\end{tabular}
\caption{Description of 5 early warning indicators (5EWSI). Time series $X$ containing $N(=400)$ elements labeled $x_i$. Here $\bar{x}$ denotes the mean of $x$ and $y_i$ are the elements of a time series that is generated by shifting $X$ by 1, so to that $y_i = x_{i-1}$.}
\label{5S}
\end{table}

\section{Performance on withheld testing set}
\begin{table}[H]
\centering
\begin{tabular}{lcrrrr}
\toprule
\multirow{2}{*}{\textbf{Evaluation}} & \multirow{2}{*}{\textbf{Model}} & \multicolumn{4}{c}{\textbf{22SF}}\\
\cmidrule(r){3-6} 
 & & \makecell[c]{WhiteN} & \makecell[c]{EnvN} & \makecell[c]{DemN}& \makecell[c]{MixedN} \\
\midrule
\multirow{4}{*}{\textbf{Accuracy}} & GBM & 0.9963 ($\pm$0.0024) & 0.9921 ($\pm$0.0035) & 0.9975($\pm$0.0020) & 0.9958 ($\pm$0.0026) \\
& LRM & 0.9963 ($\pm$0.0024) & 0.9917 ($\pm$0.0036) & 0.9979 ($\pm$0.0018) & 0.9933 ($\pm$0.0032)\\
& KNN & 0.9954 ($\pm$0.0027) & 0.9904 ($\pm$0.0039) & 0.9992 ($\pm$0.0011) & 0.9971 ($\pm$0.0022) \\
& SVM & 0.9971 ($\pm$0.0022) & 0.9917 ($\pm$0.0036) & 0.9979 ($\pm$0.0018) & 0.9938 ($\pm$0.0031) \\
\midrule
\multirow{4}{*}{\textbf{AUC}} & GBM & 0.9999 ($\pm$0.0004) & 0.9991 ($\pm$0.0012) & 1 ($\pm$0.0000) & 0.9998 ($\pm$0.0006) \\
& LRM & 1 ($\pm$0.0000) & 0.9993 ($\pm$0.0011) & 1 ($\pm$0.0000) & 0.9996 ($\pm$0.0008) \\
& KNN & 0.9995 ($\pm$0.0009) & 0.9956 ($\pm$0.0027) & 1 ($\pm$0.0000) & 0.9995 ($\pm$0.0009) \\
& SVM & 1 ($\pm$0.0000) & 0.9993 ($\pm$0.0011) & 1 ($\pm$0.0000) & 0.9995 ($\pm$0.0009) \\
\midrule
\multirow{2}{*}{\textbf{Evaluation}} & \multirow{2}{*}{\textbf{Model}} & \multicolumn{4}{c}{\textbf{5EWSI}}\\
\cmidrule(r){3-6} 
 & & \makecell[c]{WhiteN} & \makecell[c]{EnvN} & \makecell[c]{DemN} & \makecell[c]{MixedN} \\
\midrule
\multirow{4}{*}{\textbf{Accuracy}} & GBM & 0.9988 ($\pm$0.0014) & 0.9650 ($\pm$0.0074) & 0.9996 ($\pm$0.0008) & 0.9796 ($\pm$0.0057) \\
& LRM & 0.9992 ($\pm$0.0011) & 0.9604 ($\pm$0.0078) & 1 ($\pm$0.0000) & 0.9775 ($\pm$0.0059) \\
& KNN & 0.9983 ($\pm$0.0016) & 0.9642 ($\pm$0.0074) & 0.9996 ($\pm$0.0008) & 0.9825 ($\pm$0.0052) \\
& SVM & 1 ($\pm$0.0000) & 0.9621 ($\pm$0.0076) & 1 ($\pm$0.0000) & 0.9783 ($\pm$0.0058) \\
\midrule
\multirow{4}{*}{\textbf{AUC}} & GBM & 1 ($\pm$0.0000) & 0.9940 ($\pm$0.0031) & 1 ($\pm$0.0000) & 0.9983 ($\pm$0.0017) \\
& LRM & 1 ($\pm$0.0000) & 0.9938 ($\pm$0.0032) & 1 ($\pm$0.0000) & 0.9973 ($\pm$0.0021) \\
& KNN & 0.9992 ($\pm$0.0011) & 0.9911 ($\pm$0.0038) & 1 ($\pm$0.0000) & 0.9974 ($\pm$0.0020) \\
& SVM & 0.9996 ($\pm$0.0008) & 0.9940 ($\pm$0.0031) & 1 ($\pm$0.0000) & 0.9974 ($\pm$0.0020) \\
\bottomrule
\end{tabular}
\caption{Classification performance of 32 trained classifiers on withheld testing sets. 22SF, 22 statistical features; 5EWSI, 5 early warning signal indicators; WhiteN (EnvN, DemN, MixedN), simulated data from SIR model with white noise (multiplicative environmental noise, demographic noise, mixed data from the previous three datasets); AUC, rea-under-the-curve; GBM, gradient boosting machine; KNN, k-nearest neighbor; SVM, support vector machines; and LRM, logistic regression model.}
\label{SythResult}
\end{table}

\section{DeLong test results}
\subsection{Predictive model is fixed}
The DeLong test \cite{delong1988comparing} is utilized to compare the difference between the AUC scores of two classifiers. Here, we perform DeLong tests on classifiers derived from eight datasets (i.e. every unique combination of four sets of time series and two feature extraction libraries) with one predetermined predictive model. 

When using the same predictive model, AUC scores of classifiers trained from 5EWSI of MixedN (i.e., $M5G$, $M5L$, $M5K$, or $M5S$) are significantly different from those trained from other datasets ($p<0.05$), with three pairs showing statistically similar AUC scores ($p\leq0.05$): $M5G$ and $E22G$ ($p=0.1691$), $M5K$ and $E22K$ ($p=0.2702$), and $M5K$ and $W5K$ ($p=0.1125$). Additionally, classifiers trained from 22SF of EnvN (i.e., $E22G$, $E22L$, $E22K$, or $E22S$) exhibit significantly different AUC than classifiers trained from other data across four predictive models ($p<0.05$ for most cases). There are five exceptions: $M22G$ and $E22G$ ($p=0.0953$), $M5G$ and $E22G$ ($p=0.1691$), $M22L$ and $E22L$ ($p=0.5424$), $M5K$ and $E22K$ ($p=0.2702$), and $M22S$ and $E22S$ ($p=0.5320$). The remaining pairs of AUC scores demonstrate statistically identical values, indicating the same performances of their corresponding classifiers.

\begin{table}[H]
  \centering
  \begin{tabular}{lccccccccc}
    \toprule
    \multirow{2}{*}{\textbf{}} & \multirow{2}{*}{\textbf{}} & \multicolumn{8}{c}{\textbf{Classifiers}}\\
    \cmidrule(r){3-10}
    & & W22G & E22G & D22G & M22G & W5G & E5G & D5G & M5G \\
    \midrule
    \multirow{8}{*}{\textbf{Classifiers}} & W22G & \cellcolor{green!20}1 & & & & & & & \\
    & E22G & \cellcolor{yellow!20}0.0325 & \cellcolor{green!20}1 & & & & & & \\
    & D22G & \cellcolor{green!20}0.0951 & \cellcolor{yellow!20}0.0187 & \cellcolor{green!20}1 & & & & & \\
    & M22G & \cellcolor{green!20}0.4003 & \cellcolor{green!20}0.0953 & \cellcolor{green!20}0.1714 & \cellcolor{green!20}1 & & & & \\
    & W5G & \cellcolor{green!20}0.0644 & \cellcolor{yellow!20}0.0184 & \cellcolor{green!20}0.8616 & \cellcolor{green!20}0.1671 & \cellcolor{green!20}1 & & & \\
    & E5G & \cellcolor{red!20}$<$0.0001 & \cellcolor{red!20}$<$0.0001 & \cellcolor{red!20}$<$0.0001 & \cellcolor{red!20}$<$0.0001 & \cellcolor{red!20}$<$0.0001 & \cellcolor{green!20}1 & & \\
    & D5G & \cellcolor{yellow!20}0.0474 & \cellcolor{yellow!20}0.0170 & \cellcolor{green!20}0.1464 & \cellcolor{green!20}0.1466 & \cellcolor{green!20}0.2077 & \cellcolor{red!20}$<$0.0001 & \cellcolor{green!20}1 & \\
    & M5G & \cellcolor{red!20}0.0004 & \cellcolor{green!20}0.1691 & \cellcolor{red!20}0.0002 & \cellcolor{red!20}$<$0.0001 & \cellcolor{red!20}0.0002 & \cellcolor{red!20}0.0009 & \cellcolor{red!20}0.0002 & \cellcolor{green!20}1 \\
    \bottomrule
  \end{tabular}
  \caption{DeLong Test p-vaLues Results for GBM. p $<$ 0.001 (red) indicating very strong evidence against $H_0$, p $<$ 0.01 (orange) strong evidence, p $<$ 0.05 (yellow) moderate evidence, p $\geq$ 0.05 (green) not significant.}
  \label{DeLong_GBM}
\end{table}

\begin{table}[H]
\centering
\begin{tabular}{lccccccccc}
\toprule
 \multirow{2}{*}{\textbf{}} & \multirow{2}{*}{\textbf{}} & \multicolumn{8}{c}{\textbf{Classifiers}}\\
    \cmidrule(r){3-10}
    & & W22L & E22L & D22L & M22L & W5L & E5L & D5L & M5L \\
\midrule
\multirow{8}{*}{\textbf{Classifiers}} & W22L & \cellcolor{green!20}1 &  &  &  &  &  &  &  \\
& E22L & \cellcolor{yellow!20}0.0311 & \cellcolor{green!20}1 &  &  &  &  &  &  \\
& D22L & \cellcolor{green!20}0.2187 & \cellcolor{yellow!20}0.0240 & \cellcolor{green!20}1 &  &  &  &  &  \\
& M22L & \cellcolor{green!20}0.1892 & \cellcolor{green!20}0.5424 & \cellcolor{green!20}0.1580 & \cellcolor{green!20}1 &  &  &  &  \\
& W5L & \cellcolor{green!20}0.1133 & \cellcolor{yellow!20}0.0227 & \cellcolor{green!20}0.3568 & \cellcolor{green!20}0.1514 & \cellcolor{green!20}1 &  &  &  \\
& E5L & \cellcolor{red!20}$<$0.0001 & \cellcolor{red!20}$<$0.0001 & \cellcolor{red!20}$<$0.0001 & \cellcolor{red!20}$<$0.0001 & \cellcolor{red!20}$<$0.0001 & \cellcolor{green!20}1 &  &  \\
& D5L & \cellcolor{green!20}0.0858 & \cellcolor{yellow!20}0.022 & \cellcolor{green!20}0.1653 & \cellcolor{green!20}0.1488 & \cellcolor{green!20}0.3171 & \cellcolor{red!20}$<$0.0001 & \cellcolor{green!20}1 &  \\
& M5L & \cellcolor{red!20}0.0004 & \cellcolor{yellow!20}0.0146 & \cellcolor{red!20}0.0004 & \cellcolor{yellow!20}0.0056 & \cellcolor{red!20}0.0004 & \cellcolor{orange!50}0.0059 & \cellcolor{red!20}0.0004 & \cellcolor{green!20}1 \\
\bottomrule
\end{tabular}
\caption{DeLong Test p-vaLues Results for LRM. p $<$ 0.001 (red) indicating very strong evidence against $H_0$, p $<$ 0.01 (orange) strong evidence, p $<$ 0.05 (yellow) moderate evidence, p $\geq$ 0.05 (green) not significant.}
\label{DeLong_LRM}
\end{table}

\begin{table}[H]
\centering
\begin{tabular}{lccccccccc}
\toprule
 \multirow{2}{*}{\textbf{}} & \multirow{2}{*}{\textbf{}} & \multicolumn{8}{c}{\textbf{Classifiers}}\\
    \cmidrule(r){3-10}
    & & W22K & E22K & D22K & M22K & W5K & E5K & D5K & M5K \\
\midrule
\multirow{8}{*}{\textbf{Classifiers}} & W22K & \cellcolor{green!20}1 &  &  &  &  &  &  &  \\
& E22K & \cellcolor{orange!50}0.0064 & \cellcolor{green!20}1 &  &  &  &  &  &  \\
& D22K & \cellcolor{green!20}0.2617 & \cellcolor{orange!50}0.0013 & \cellcolor{green!20}1 &  &  &  &  &  \\
& M22K & \cellcolor{green!20}0.9666 & \cellcolor{orange!50}0.0061 & \cellcolor{green!20}0.2841 & \cellcolor{green!20}1 &  &  &  &  \\
& W5K & \cellcolor{green!20}0.6366 & \cellcolor{yellow!20}0.0169 & \cellcolor{green!20}0.1582 & \cellcolor{green!20}0.6113 & \cellcolor{green!20}1 &  &  &  \\
& E5K & \cellcolor{red!20}$<$0.0001 & \cellcolor{yellow!20}0.0116 & \cellcolor{red!20}$<$0.0001 & \cellcolor{red!20}$<$0.0001 & \cellcolor{red!20}$<$0.0001 & \cellcolor{green!20}1 &  &  \\
& D5K & \cellcolor{green!20}0.2570 & \cellcolor{orange!50}0.0012 & \cellcolor{green!20}0.2527 & \cellcolor{green!20}0.2790 & \cellcolor{green!20}0.1558 & \cellcolor{red!20}$<$0.0001 & \cellcolor{green!20}1 &  \\
& M5K & \cellcolor{yellow!20}0.0396 & \cellcolor{green!20}0.2702 & \cellcolor{orange!50}0.0051 & \cellcolor{orange!50}0.0085 & \cellcolor{green!20}0.1125 & \cellcolor{orange!50}0.0012 & \cellcolor{orange!50}0.0050 & \cellcolor{green!20}1 \\
\bottomrule
\end{tabular}
\caption{DeLong Test p-vaLues Results for KNN. p $<$ 0.001 (red) indicating very strong evidence against $H_0$, p $<$ 0.01 (orange) strong evidence, p $<$ 0.05 (yellow) moderate evidence, p $\geq$ 0.05 (green) not significant.}
\label{DeLong_KNN}
\end{table}

\begin{table}[H]
\centering
\begin{tabular}{lccccccccc}
\toprule
 \multirow{2}{*}{\textbf{}} & \multirow{2}{*}{\textbf{}} & \multicolumn{8}{c}{\textbf{Classifiers}}\\
    \cmidrule(r){3-10}
    & & W22S & E22S & D22S & M22S & W5S & E5S & D5S & M5S \\
\midrule
\multirow{8}{*}{\textbf{Classifiers}} & W22S & \cellcolor{green!20}1 &  &  &  &  &  &  &  \\
& E22S & \cellcolor{yellow!20}0.0216 & \cellcolor{green!20}1 &  &  &  &  &  &  \\
& D22S & \cellcolor{green!20}0.1567 & \cellcolor{yellow!20}0.0157 & \cellcolor{green!20}1 &  &  &  &  &  \\
& M22S & \cellcolor{green!20}0.2181 & \cellcolor{green!20}0.5320 & \cellcolor{green!20}0.1819 & \cellcolor{green!20}1 &  &  &  &  \\
& W5S & \cellcolor{green!20}0.9879 & \cellcolor{yellow!20}0.0222 & \cellcolor{green!20}0.4076 & \cellcolor{green!20}0.2196 & \cellcolor{green!20}1 &  &  &  \\
& E5S & \cellcolor{red!20}$<$0.0001 & \cellcolor{red!20}$<$0.0001 & \cellcolor{red!20}$<$0.0001 & \cellcolor{red!20}$<$0.0001 & \cellcolor{red!20}$<$0.0001 & \cellcolor{green!20}1 &  &  \\
& D5S & \cellcolor{green!20}0.0860 & \cellcolor{yellow!20}0.0148 & \cellcolor{green!20}0.2203 & \cellcolor{green!20}0.1760 & \cellcolor{green!20}0.3213 & \cellcolor{red!20}$<$0.0001 & \cellcolor{green!20}1 &  \\
& M5S & \cellcolor{red!20}0.0004 & \cellcolor{yellow!20}0.0187 & \cellcolor{red!20}0.0004 & \cellcolor{orange!50}0.0069 & \cellcolor{red!20}$<$0.0001 & \cellcolor{orange!50}0.0065 & \cellcolor{red!20}0.0004 & \cellcolor{green!20}1 \\
\bottomrule
\end{tabular}
\caption{DeLong Test p-vaLues Results for SVM. p $<$ 0.001 (red) indicating very strong evidence against $H_0$, p $<$ 0.01 (orange) strong evidence, p $<$ 0.05 (yellow) moderate evidence, p $\geq$ 0.05 (green) not significant.}
\label{DeLong_SVM}
\end{table}

\newpage
\subsection{Training data is fixed}
Additionally, we perform DeLong tests on classifiers derived from four predictive models with eight datasets (i.e. every unique combination of four sets of time series and two feature extraction libraries). Eight tables (Table \ref{DeLong_white_22SF}-\ref{DeLong_mixed_EWSI}) are provided, each corresponding to a dataset, to depict the p-values.
\begin{table}[H]
\centering
\begin{tabular}{lccccc}
\toprule
\multirow{2}{*}{\textbf{}} & \multirow{2}{*}{\textbf{}} & \multicolumn{4}{c}{\textbf{Algorithms}} \\
\cmidrule(r){3-6}
 & & GBM & LRM & KNN & SVM \\
\midrule
\multirow{4}{*}{\textbf{Algorithms}} & GBM & \cellcolor{green!20}1 &  &  & \\
& LRM & \cellcolor{green!20}0.3232 & \cellcolor{green!20}1 &  &  \\
& KNN & \cellcolor{green!20}0.3468 & \cellcolor{green!20}0.2972 & \cellcolor{green!20}1 &  \\
& SVM & \cellcolor{green!20}0.1147 & \cellcolor{green!20}0.9660 & \cellcolor{green!20}0.2963 & \cellcolor{green!20}1 \\
\bottomrule
\end{tabular}
\caption{DeLong Test p-vaLues Results for fixed data: 22SF of White Noise data WhiteN. p $<$ 0.001 (red) indicating very strong evidence against $H_0$, p $<$ 0.01 (orange) strong evidence, p $<$ 0.05 (yellow) moderate evidence, p $\geq$ 0.05 (green) not significant.}
\label{DeLong_white_22SF}
\end{table}

\begin{table}[H]
\centering
\begin{tabular}{lccccc}
\toprule
\multirow{2}{*}{\textbf{}} & \multirow{2}{*}{\textbf{}} & \multicolumn{4}{c}{\textbf{Algorithms}} \\
\cmidrule(r){3-6}
 & & GBM & LRM & KNN & SVM \\
\midrule
\multirow{4}{*}{\textbf{Algorithms}} & GBM & \cellcolor{green!20}1 &  &  & \\
& LRM & \cellcolor{green!20}0.6767 & \cellcolor{green!20}1 &  &  \\
& KNN & \cellcolor{orange!50}0.0048 & \cellcolor{orange!50}0.0083 & \cellcolor{green!20}1 &  \\
& SVM & \cellcolor{green!20}0.6266 & \cellcolor{green!20}0.9282 & \cellcolor{orange!50}0.0019 & \cellcolor{green!20}1 \\
\bottomrule
\end{tabular}
\caption{DeLong Test p-vaLues Results for fixed data: 22SF of Multiplicative Environmental Noise data EnvN. p $<$ 0.001 (red) indicating very strong evidence against $H_0$, p $<$ 0.01 (orange) strong evidence, p $<$ 0.05 (yellow) moderate evidence, p $\geq$ 0.05 (green) not significant.}
\label{DeLong_env_22SF}
\end{table}

\begin{table}[H]
\centering
\begin{tabular}{lccccc}
\toprule
\multirow{2}{*}{\textbf{}} & \multirow{2}{*}{\textbf{}} & \multicolumn{4}{c}{\textbf{Algorithms}} \\
\cmidrule(r){3-6}
 & & GBM & LRM & KNN & SVM \\
\midrule
\multirow{4}{*}{\textbf{Algorithms}} & GBM & \cellcolor{green!20}1 &  &  & \\
& LRM & \cellcolor{green!20}0.8024 & \cellcolor{green!20}1 &  &  \\
& KNN & \cellcolor{green!20}0.2698 & \cellcolor{green!20}0.5530 & \cellcolor{green!20}1 &  \\
& SVM & \cellcolor{green!20}0.3819 & \cellcolor{green!20}0.6886 & \cellcolor{green!20}0.7441 & \cellcolor{green!20}1 \\
\bottomrule
\end{tabular}
\caption{DeLong Test p-vaLues Results for fixed data: 22SF of Demographic Noise data DemN. p $<$ 0.001 (red) indicating very strong evidence against $H_0$, p $<$ 0.01 (orange) strong evidence, p $<$ 0.05 (yellow) moderate evidence, p $\geq$ 0.05 (green) not significant.}
\label{DeLong_dem_22SF}
\end{table}

\begin{table}[H]
\centering
\begin{tabular}{lccccc}
\toprule
\multirow{2}{*}{\textbf{}} & \multirow{2}{*}{\textbf{}} & \multicolumn{4}{c}{\textbf{Algorithms}} \\
\cmidrule(r){3-6}
 & & GBM & LRM & KNN & SVM \\
\midrule
\multirow{4}{*}{\textbf{Algorithms}} & GBM & \cellcolor{green!20}1 &  &  & \\
& LRM & \cellcolor{green!20}0.5244 & \cellcolor{green!20}1 &  &  \\
& KNN & \cellcolor{green!20}0.3912 & \cellcolor{green!20}0.9565 & \cellcolor{green!20}1 &  \\
& SVM & \cellcolor{green!20}0.2209 & \cellcolor{green!20}0.9683 & \cellcolor{green!20}0.9207 & \cellcolor{green!20}1 \\
\bottomrule
\end{tabular}
\caption{DeLong Test p-vaLues Results for fixed data: 22SF of Mixed Noise data MixedN. p $<$ 0.001 (red) indicating very strong evidence against $H_0$, p $<$ 0.01 (orange) strong evidence, p $<$ 0.05 (yellow) moderate evidence, p $\geq$ 0.05 (green) not significant.}
\label{DeLong_mixed_22SF}
\end{table}

\begin{table}[H]
\centering
\begin{tabular}{lccccc}
\toprule
\multirow{2}{*}{\textbf{}} & \multirow{2}{*}{\textbf{}} & \multicolumn{4}{c}{\textbf{Algorithms}} \\
\cmidrule(r){3-6}
 & & GBM & LRM & KNN & SVM \\
\midrule
\multirow{4}{*}{\textbf{Algorithms}} & GBM & \cellcolor{green!20}1 &  &  & \\
& LRM & \cellcolor{green!20}0.3307 & \cellcolor{green!20}1 &  &  \\
& KNN & \cellcolor{green!20}0.1575 & \cellcolor{green!20}0.1561 & \cellcolor{green!20}1 &  \\
& SVM & \cellcolor{green!20}0.4571 & \cellcolor{green!20}0.3308 & \cellcolor{green!20}0.1559 & \cellcolor{green!20}1 \\
\bottomrule
\end{tabular}
\caption{DeLong Test p-vaLues Results for fixed data: 5EWSI of White Noise data WhiteN. p $<$ 0.001 (red) indicating very strong evidence against $H_0$, p $<$ 0.01 (orange) strong evidence, p $<$ 0.05 (yellow) moderate evidence, p $\geq$ 0.05 (green) not significant.}
\label{DeLong_white_EWSI}
\end{table}

\begin{table}[H]
\centering
\begin{tabular}{lccccc}
\toprule
\multirow{2}{*}{\textbf{}} & \multirow{2}{*}{\textbf{}} & \multicolumn{4}{c}{\textbf{Algorithms}} \\
\cmidrule(r){3-6}
 & & GBM & LRM & KNN & SVM \\
\midrule
\multirow{4}{*}{\textbf{Algorithms}} & GBM & \cellcolor{green!20}1 &  &  & \\
& LRM & \cellcolor{green!20}0.8203 & \cellcolor{green!20}1 &  &  \\
& KNN & \cellcolor{yellow!20}0.0186 & \cellcolor{yellow!20}0.0294 & \cellcolor{green!20}1 &  \\
& SVM & \cellcolor{green!20}0.9857 & \cellcolor{green!20}0.2537 & \cellcolor{yellow!20}0.01674 & \cellcolor{green!20}1 \\
\bottomrule
\end{tabular}
\caption{DeLong Test p-vaLues Results for fixed data: 5EWSI of Multiplicative Environmental Noise data EnvN. p $<$ 0.001 (red) indicating very strong evidence against $H_0$, p $<$ 0.01 (orange) strong evidence, p $<$ 0.05 (yellow) moderate evidence, p $\geq$ 0.05 (green) not significant.}
\label{DeLong_env_EWSI}
\end{table}

\begin{table}[H]
\centering
\begin{tabular}{lccccc}
\toprule
\multirow{2}{*}{\textbf{}} & \multirow{2}{*}{\textbf{}} & \multicolumn{4}{c}{\textbf{Algorithms}} \\
\cmidrule(r){3-6}
 & & GBM & LRM & KNN & SVM \\
\midrule
\multirow{4}{*}{\textbf{Algorithms}} & GBM & \cellcolor{green!20}1 &  &  & \\
& LRM & \cellcolor{green!20}1 & \cellcolor{green!20}1 &  &  \\
& KNN & \cellcolor{green!20}1 & \cellcolor{green!20}1 & \cellcolor{green!20}1 &  \\
& SVM & \cellcolor{green!20}1 & \cellcolor{green!20}1 & \cellcolor{green!20}1 & \cellcolor{green!20}1 \\
\bottomrule
\end{tabular}
\caption{DeLong Test p-vaLues Results for fixed data: 5EWSI of Demographic Noise data DemN. Note that DeLong Test of two ROC curves with AUC == 1 has always p.value = 1 and can be misleading. p $<$ 0.001 (red) indicating very strong evidence against $H_0$, p $<$ 0.01 (orange) strong evidence, p $<$ 0.05 (yellow) moderate evidence, p $\geq$ 0.05 (green) not significant.}
\label{DeLong_dem_EWSI}
\end{table}

\begin{table}[H]
\centering
\begin{tabular}{lccccc}
\toprule
\multirow{2}{*}{\textbf{}} & \multirow{2}{*}{\textbf{}} & \multicolumn{4}{c}{\textbf{Algorithms}} \\
\cmidrule(r){3-6}
 & & GBM & LRM & KNN & SVM \\
\midrule
\multirow{4}{*}{\textbf{Algorithms}} & GBM & \cellcolor{green!20}1 &  &  & \\
& LRM & \cellcolor{yellow!20}0.0174 & \cellcolor{green!20}1 &  &  \\
& KNN & \cellcolor{green!20}0.2316 & \cellcolor{green!20}0.8783 & \cellcolor{green!20}1 &  \\
& SVM & \cellcolor{yellow!20}0.0230 & \cellcolor{green!20}0.2333 & \cellcolor{green!20}0.9239 & \cellcolor{green!20}1 \\
\bottomrule
\end{tabular}
\caption{DeLong Test p-vaLues Results for fixed data: 5EWSI of Mixed Noise data MixedN. p $<$ 0.001 (red) indicating very strong evidence against $H_0$, p $<$ 0.01 (orange) strong evidence, p $<$ 0.05 (yellow) moderate evidence, p $\geq$ 0.05 (green) not significant.}
\label{DeLong_mixed_EWSI}
\end{table}

\newpage
\section{Classification accuracy results of Rolling window and Expanding window experiments}
\begin{figure}[H]
    \centering
    \includegraphics[width=\textwidth]{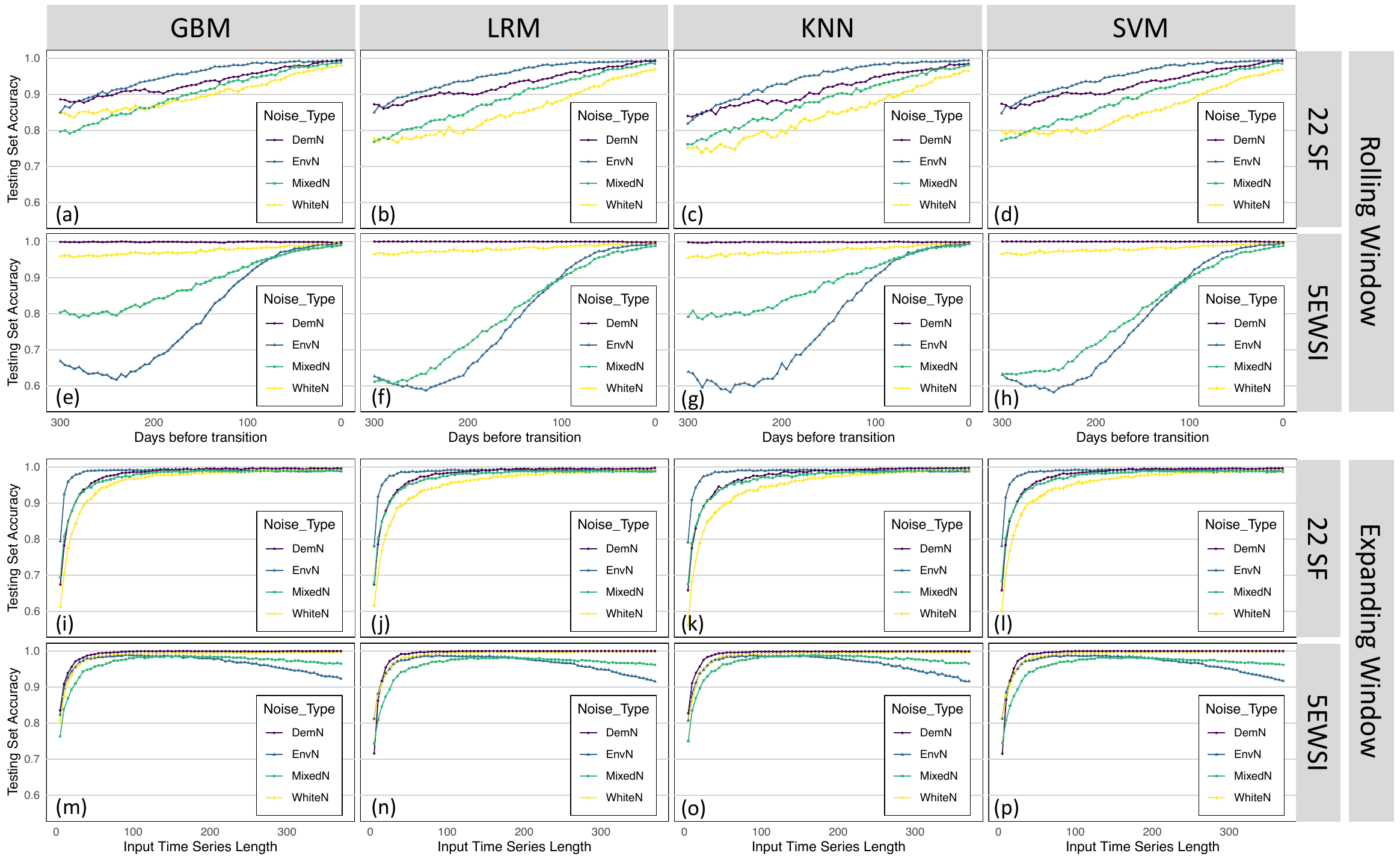}
    \caption{Accuracy of synthetic testing sets. We recorded the results of both rolling window and expanding window approaches across 8 datasets and 4 predictive models.}
    \label{EarlinessACC}
\end{figure}

\section{Information of COVID-19 incidence data}
\label{I18}
The SIR model assumes the population is homogeneously mixed, meaning the contact probability between two individuals is identical. However, this assumption may not hold in most country-level studies. Nevertheless, the national-level data is common and relatively accurate owing to its broader research base. We additionally gather the COVID-19 daily incidence data across 18 countries from “Our World in Data”~\cite{owidcoronavirus} for the case study.

\textbf{$I_{18}$ (\# replicates = 194; COVID-19 incidence data from 18 countries)} To derive the replicates and the corresponding labels, we apply the $R_e$ estimation from Huisman et al.~\cite{huisman2022estimation} for COVID-19 incidence data, where the assumption is that the incidence data come from symptomatic individuals. The estimation is unbiased if the proportion of asymptomatic or presymptomatic cases remains constant. Given the typical lag between disease occurrence and data collection, subsequences with lengths shorter than 14 days (i.e., two weeks), or replicates containing missing data are excluded. The final dataset comprises 194 replicates, labeled as “T”, encompassing data from 18 countries. Detailed information on the sliced data is provided in Supplementary Table~\ref{countries}.

We should acknowledge that the SIR model does not consider the asymptomatic individuals, whereas presymptomatic and asymptomatic cases take more than half of total infection at the outbreak peak, resulting in highly underestimated incidence data~\cite{raghunath2020prediction}. Despite this inherent limitation, our classification approach remains robust and barely affected by the underestimated cases.

For $I_{18}$, results suggest that classifiers trained from the KNN model outperform those trained by the other three models, among which $E22K$ has the lowest accuracy $0.7990$ $\pm 0.0564$. Classifiers trained from 5EWSI and KNN (i.e. W5K, E5K, D5K, and M5K) have the accuracy of $1$ $\pm 0.0000$, despite the original training time series being used. For $I_{SG}$, classifiers trained from KNN have high accuracy. Four classifiers (E5G, E5L, E5K, and E5S) that are trained from 5EWSI of $I_E$ also exhibit satisfying performance, with the accuracy ranging from $0.7629$ $\pm0.0593$ of E5G to $1$ $\pm 0.0000$ of the rest. Classifiers achieved better performance on $I_{SG}$ compared to $I_{18}$, which is consistent with our rationale that Singapore data is more suitable for SIR model assumptions.

\begin{table}[H]
  \centering
  \begin{tabular}{ccccc}
    \toprule
    \textbf{ISO Code} & \textbf{Country} & \textbf{Continent} & \makecell{\textbf{Number of} \\ \textbf{Sliced Sequence}} & \makecell{\textbf{Range of }\\ \textbf{Sequence Length}} \\
    \midrule
    \text{ARE} & \text{United Arab Emirates} & \text{Asia} & 13 & 16-127\\
    \text{BGD} & \text{Bangladesh} & \text{Asia} & 11 & 14-98 \\
    \text{BRA} & \text{Brazil} & \text{South America} & 11 & 14-39 \\
    \text{CAN} & \text{Canada} & \text{North America} & 7 & 16-94 \\
    \text{CHL} & \text{Chile} & \text{South America} & 12 & 18-97 \\
    \text{COL} & \text{Colombia} & \text{South America} & 12 & 14-112 \\
    \text{EGY} & \text{Egypt} & \text{Africa} & 10 & 14-73 \\
    \text{FIN} & \text{Finland} & \text{Europe} & 11 & 14-69 \\
    \text{IND} & \text{India} & \text{Asia} & 10 & 16-149 \\
    \text{JPN} & \text{Japan} & \text{Asia} & 10 & 22-96 \\
    \text{KOR} & \text{South Korea} & \text{Asia} & 10 & 16-96\\
    \text{MEX} & \text{Mexico} & \text{North America} & 11 & 14-99 \\
    \text{MYS} & \text{Malaysia} & \text{Asia} & 13 & 14-74 \\
    \text{NGA} & \text{Nigeria} & \text{Africa} & 16 & 14-88 \\
    \text{NOR} & \text{Norway} & \text{Europe} & 11 & 14-103 \\
    \text{SWE} & \text{Sweden} & \text{Europe} & 9 & 14-87\\
    \text{THA} & \text{Thailand} & \text{Asia} & 10 & 14-136 \\
    \text{USA} & \text{United States} & \text{North America} & 7 & 16-67 \\
    \bottomrule
  \end{tabular}
  \caption{ISO Codes and Corresponding Countries with Continents. Sequences containing missing data, or sequence length shorter than 14 are eliminated. As a result, the COVID-19 from 18 countries dataset consists of 194 samples from the original data, with sample lengths ranging from 14 to 149.}
  \label{countries}
\end{table}

\section{Effective reproduction number estimation of SARS data}
\begin{figure}[H]
    \centering
    \includegraphics[width=\textwidth]{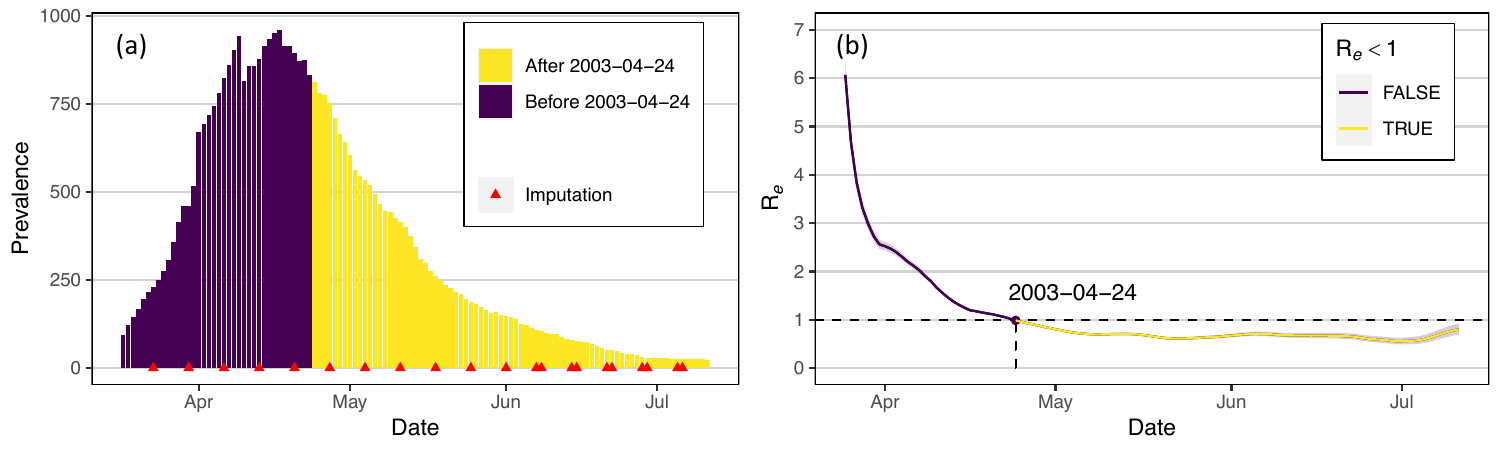}
    \caption{(a) SARS prevalence data of Hong Kong. The missing data of the original sequence is imputed using \texttt{imputeTS}. (b) \textbf{$R_e$} estimation results through \texttt{EpiEstim} with a mean value of 8.4 and standard deviation of 3.8. The shaded region is the $95\%$ confidence interval.}
    \label{HK_Data_Re}
\end{figure}

\section{Performance on empirical testing sets}
\begin{table}[H]
\centering
\begin{tabular}{lcrrrr}
\toprule
\multirow{2}{*}{\textbf{Evaluation}} & \multirow{2}{*}{\textbf{Model}} & \multicolumn{4}{c}{\textbf{22SF}}\\
\cmidrule(r){3-6} 
& & \makecell[c]{WhiteN} & \makecell[c]{EnvN} & \makecell[c]{DemN}& \makecell[c]{MixedN} \\
\midrule
\multirow{4}{*}{\textbf{Accuracy}} & GBM & 0.6392 ($\pm$0.0676) & 0.3608 ($\pm$0.0676) & 0.1701 ($\pm$0.0529) & 0.5258 ($\pm$0.0703) \\
& LRM & 0 ($\pm$0.0000) & 0.0258 ($\pm$0.0223) & 0.1598 ($\pm$0.0516) & 0.2010 ($\pm$0.0564) \\
& KNN & 0.9588 ($\pm$0.0280) & 0.7990 ($\pm$0.0564) & 0.8454 ($\pm$0.0509) & 0.8454 ($\pm$0.0509) \\
& SVM & 0 ($\pm$0.0000) & 0.0052 ($\pm$0.0101) & 0.1907 ($\pm$0.0553) & 0.1495 ($\pm$0.0502) \\
\midrule
\multirow{2}{*}{\textbf{Evaluation}} & \multirow{2}{*}{\textbf{Model}} & \multicolumn{4}{c}{\textbf{5EWSI}}\\
\cmidrule(r){3-6} 
& & \makecell[c]{WhiteN} & \makecell[c]{EnvN} & \makecell[c]{DemN}& \makecell[c]{MixedN} \\
\midrule
\multirow{4}{*}{\textbf{Accuracy}} & GBM & 0.5619 ($\pm$0.0698) & 0.8041 ($\pm$0.0559) & 0.5670 ($\pm$0.0697) & 0.7371 ($\pm$0.0619) \\
& LRM & 1 ($\pm$0.0000) & 1 ($\pm$0.0000) & 0 ($\pm$0.0000) & 0.0103 ($\pm$0.0142) \\
& KNN & 1 ($\pm$0.0000) & 1 ($\pm$0.0000) & 1 ($\pm$0.0000) & 1 ($\pm$0.0000) \\
& SVM & 1 ($\pm$0.0000) & 1 ($\pm$0.0000) & 0 ($\pm$0.0000) & 0.0103 ($\pm$0.0142) \\
\bottomrule
\end{tabular}
\caption{Classification accuracy of 32 synthetic-data-trained classifiers for COVID-19 Incidence Data of 18 Countries or Regions.}
\label{18CovidResult}
\end{table}

\begin{table}[H]
\centering
\begin{tabular}{lcrrrr}
\toprule
\multirow{2}{*}{\textbf{Evaluation}} & \multirow{2}{*}{\textbf{Model}} & \multicolumn{4}{c}{\textbf{22SF}}\\
\cmidrule(r){3-6}
& & \makecell[c]{WhiteN} & \makecell[c]{EnvN} & \makecell[c]{DemN}& \makecell[c]{MixedN} \\
\midrule
\multirow{4}{*}{\textbf{Accuracy}} & GBM &  0.8947($\pm$0.1380) &  0.5789($\pm$0.2220) &  0.0526($\pm$0.1004) &  0.8947($\pm$0.1380) \\
& LRM &  0($\pm$0.0000) &  0($\pm$0.0000) &  0.2105($\pm$0.1833) & 0.1053($\pm$0.1380) \\
& KNN &  1($\pm$0.0000) &  1($\pm$0.0000) & 1($\pm$0.0000) & 1($\pm$0.0000) \\
& SVM &  0.0526($\pm$0.1004) &  0($\pm$0.0000) &  0.3158($\pm$0.2090) &  0.1053($\pm$0.1380) \\
\midrule
\multirow{2}{*}{\textbf{Evaluation}} & \multirow{2}{*}{\textbf{Model}} &  \multicolumn{4}{c}{\textbf{5EWSI}}\\
\cmidrule(r){3-6} 
& & \makecell[c]{WhiteN} & \makecell[c]{EnvN} & \makecell[c]{DemN}& \makecell[c]{MixedN} \\
\midrule
\multirow{4}{*}{\textbf{Accuracy}} & GBM & 1($\pm$0.0000) &  0.9474($\pm$0.1004) &  0.0526($\pm$0.1004) &  0.9474($\pm$0.1004) \\
& LRM &  1($\pm$0.0000) &  1($\pm$0.0000) &  0($\pm$0.0000) & 0.2105($\pm$0.1833) \\
& KNN &  1($\pm$0.0000) &  1($\pm$0.0000) &  0.9474($\pm$0.1004) &  1($\pm$0.0000) \\
& SVM &  1($\pm$0.0000) &  1($\pm$0.0000) &  0($\pm$0.0000) &  0.2105($\pm$0.1833) \\
\bottomrule
\end{tabular}
\caption{Classification accuracy of 32 synthetic-data-trained classifiers for COVID-19 Incidence Data of Singapore.}
\label{Singapore}
\end{table}

\begin{table}[H]
\centering
\begin{tabular}{lcrrrr}
\toprule
\multirow{2}{*}{\textbf{Evaluation}} & \multirow{2}{*}{\textbf{Model}} & \multicolumn{4}{c}{\textbf{22SF}}\\
\cmidrule(r){3-6}
& & \makecell[c]{WhiteN} & \makecell[c]{EnvN} & \makecell[c]{DemN}& \makecell[c]{MixedN} \\
\midrule
\multirow{4}{*}{\textbf{Accuracy}} & GBM & 0.2900 ($\pm$0.0257) & 0.1542 ($\pm$0.0204) & 1 ($\pm$0.0000) & 0.0750 ($\pm$0.0149) \\
& LRM & 1 ($\pm$0.0000) & 0.3792 ($\pm$0.0275) & 0.7267 ($\pm$0.0252) & 0.9225 ($\pm$0.0151) \\
& KNN & 0 ($\pm$0.0000) & 0 ($\pm$0.0000) & 0 ($\pm$0.0000) & 0 ($\pm$0.0000) \\
& SVM & 1 ($\pm$0.0000) & 0.7525 ($\pm$0.0244) & 0.2075 ($\pm$0.0229) & 0.0967 ($\pm$0.0167)\\
\midrule
\multirow{2}{*}{\textbf{Evaluation}} & \multirow{2}{*}{\textbf{Model}} &  \multicolumn{4}{c}{\textbf{5EWSI}}\\
\cmidrule(r){3-6} 
& & \makecell[c]{WhiteN} & \makecell[c]{EnvN} & \makecell[c]{DemN}& \makecell[c]{MixedN} \\
\midrule
\multirow{4}{*}{\textbf{Accuracy}} & GBM & 0.0400 ($\pm$0.0111) & 0 ($\pm$0.0000) & 0.7217 ($\pm$0.0254) & 0 ($\pm$0.0000) \\
& LRM & 1 ($\pm$0.0000) & 1 ($\pm$0.0000) & 1 ($\pm$0.0000) & 0.1575 ($\pm$0.0206) \\
& KNN & 0 ($\pm$0.0000) & 0 ($\pm$0.0000) & 0 ($\pm$0.0000) & 0 ($\pm$0.0000) \\
& SVM & 0 ($\pm$0.0000) & 0 ($\pm$0.0000) & 1 ($\pm$0.0000) & 0.8517 ($\pm$0.0201)\\
\bottomrule
\end{tabular}
\caption{Classification accuracy of 32 synthetic-data-trained classifiers for SARS 2003 Data in Hong Kong.}
\label{SARSresult}
\end{table}

\subsection{Performance on shorter empirical testing sets}
Implementation of preventative measures takes time. Hence, we assess the prediction accuracy by allowing one week for preparation. This is achieved by excluding seven data points right before the transition point for all empirical testing samples of COVID-19 data. Results are presented in Table \ref{18CovidResult_Shorter} and Table \ref{Singapore_Shorter}.
\begin{table}[H]
\centering
\begin{tabular}{lcrrrr}
\toprule
\multirow{2}{*}{\textbf{Evaluation}} & \multirow{2}{*}{\textbf{Model}} & \multicolumn{4}{c}{\textbf{22SF}}\\
\cmidrule(r){3-6}
& & \makecell[c]{WhiteN} & \makecell[c]{EnvN} & \makecell[c]{DemN}& \makecell[c]{MixedN} \\
\midrule
\multirow{4}{*}{\textbf{Accuracy}} & GBM & 0.6856 ($\pm$0.0653) & 0.3041 ($\pm$0.0647) & 0.1546 ($\pm$0.0509) & 0.5000 ($\pm$0.0704)\\
& LRM & 0.0103 ($\pm$0.0142) & 0.0155 ($\pm$0.0174) & 0.1495 ($\pm$0.0502) & 0.2216 ($\pm$0.0584)\\
& KNN & 0.9742 ($\pm$0.0223) & 0.8247 ($\pm$0.0535) & 0.8969 ($\pm$0.0428) & 0.8969 ($\pm$0.0428)\\
& SVM & 0.0103 ($\pm$0.0142) & 0.0052 ($\pm$0.0101) & 0.1856 ($\pm$0.0547) & 0.1598 ($\pm$0.0516)\\
\midrule
\multirow{2}{*}{\textbf{Evaluation}} & \multirow{2}{*}{\textbf{Model}} &  \multicolumn{4}{c}{\textbf{5EWSI}}\\
\cmidrule(r){3-6} 
& & \makecell[c]{WhiteN} & \makecell[c]{EnvN} & \makecell[c]{DemN}& \makecell[c]{MixedN} \\
\midrule
\multirow{4}{*}{\textbf{Accuracy}} & GBM & 0.5412 ($\pm$0.0701) & 0.7629 ($\pm$0.0593) & 0.5515 ($\pm$0.070) & 0.6907 ($\pm$0.0650)\\
& LRM & 1 ($\pm$0.0000) & 1 ($\pm$0.0000) & 0 ($\pm$0.0000) & 0.0103 ($\pm$0.0142)\\
& KNN & 1 ($\pm$0.0000) & 1 ($\pm$0.0000) & 1 ($\pm$0.0000) & 1 ($\pm$0.0000)\\
& SVM & 1 ($\pm$0.0000) & 1 ($\pm$0.0000) & 0 ($\pm$0.0000) & 0.0103 ($\pm$0.0142)\\
\bottomrule
\end{tabular}
\caption{Classification accuracy of 32 synthetic-data-trained classifiers for Shorter COVID-19 incidence data of 18 countries or regions.}
\label{18CovidResult_Shorter}
\end{table}

\begin{table}[H]
\centering
\begin{tabular}{lcrrrr}
\toprule
\multirow{2}{*}{\textbf{Evaluation}} & \multirow{2}{*}{\textbf{Model}} & \multicolumn{4}{c}{\textbf{22SF}}\\
\cmidrule(r){3-6}
& & \makecell[c]{WhiteN} & \makecell[c]{EnvN} & \makecell[c]{DemN}& \makecell[c]{MixedN} \\
\midrule
\multirow{4}{*}{\textbf{Accuracy}} & GBM & 0.8333($\pm$0.1826) &   0.3889($\pm$0.2389) & 0($\pm$0.0000) & 0.6667($\pm$0.2310) \\
& LRM & 0($\pm$0.0000) &  0($\pm$0.0000) & 0.3333($\pm$0.2310) & 0.1111($\pm$0.1540) \\
& KNN & 1($\pm$0.0000) &  1($\pm$0.0000) & 1($\pm$0.0000) & (1$\pm$0.0000) \\
& SVM & 0($\pm$0.0000) & 0 ($\pm$0.0000) & 0.2222($\pm$0.2037) & 0.0556($\pm$0.1123) \\
\midrule
\multirow{2}{*}{\textbf{Evaluation}} & \multirow{2}{*}{\textbf{Model}} &  \multicolumn{4}{c}{\textbf{5EWSI}}\\
\cmidrule(r){3-6} 
& & \makecell[c]{WhiteN} & \makecell[c]{EnvN} & \makecell[c]{DemN}& \makecell[c]{MixedN} \\
\midrule
\multirow{4}{*}{\textbf{Accuracy}} & GBM & 1($\pm$0.0000) & 1($\pm$0.0000) & 0.0625 ($\pm$0.1186) & 1($\pm$0.0000) \\
& LRM & 1($\pm$0.0000) &  1($\pm$0.0000) & 0($\pm$0.0000) & 0.2500($\pm$0.2122) \\
& KNN & 1($\pm$0.0000) & 1($\pm$0.0000) & 1($\pm$0.0000) & 1($\pm$0.0000) \\
& SVM & 1($\pm$0.0000) &  1($\pm$0.0000) & 0($\pm$0.0000) & 0.2500($\pm$0.2122) \\
\bottomrule
\end{tabular}
\caption{Classification accuracy of 32 synthetic-data-trained classifiers for Shorter COVID-19 Incidence Data of Singapore.}
\label{Singapore_Shorter}
\end{table}

\subsection{Performance on larger empirical testing sets}
COVID-19 incidence data is often significantly underestimated due to various factors, including presymptomatic and asymptomatic cases that can not be ignored. Although the estimation of the effective reproduction number ($R_e$) accounts for such underestimation, in practice, people often rely on raw incidence data that can be potentially smaller than the actual infection numbers. Here, we further test the performance of the classifiers on data expanded fivefold. Accuracy in correctly classifying “larger” testing sets is illustrated in Table \ref{18CovidResult_Larger} and Table \ref{Singapore_Larger}.
\begin{table}[H]
\centering
\begin{tabular}{lcrrrr}
\toprule
\multirow{2}{*}{\textbf{Evaluation}} & \multirow{2}{*}{\textbf{Model}} & \multicolumn{4}{c}{\textbf{22SF}}\\
\cmidrule(r){3-6}
& & \makecell[c]{WhiteN} & \makecell[c]{EnvN} & \makecell[c]{DemN}& \makecell[c]{MixedN} \\
\midrule
\multirow{4}{*}{\textbf{Accuracy}} & GBM & 0.6837($\pm$0.0651) & 0.3010($\pm$0.0642) & 0.1531($\pm$0.0504) & 0.4949($\pm$0.070)\\
& LRM & 0.0103($\pm$0.0141) & 0.0155($\pm$0.0173) & 0.1495($\pm$0.0499) & 0.2216($\pm$0.0581)\\
& KNN & 0.9742($\pm$0.0222) & 0.8247($\pm$0.0532) & 0.8969($\pm$0.0426) & 0.8969($\pm$0.0426)\\
& SVM & 0.0103($\pm$0.0141) & 0.0052($\pm$0.0101) & 0.1856($\pm$0.0544) & 0.1598($\pm$0.0513)\\
\midrule
\multirow{2}{*}{\textbf{Evaluation}} & \multirow{2}{*}{\textbf{Model}} &  \multicolumn{4}{c}{\textbf{5EWSI}}\\
\cmidrule(r){3-6} 
& & \makecell[c]{WhiteN} & \makecell[c]{EnvN} & \makecell[c]{DemN}& \makecell[c]{MixedN} \\
\midrule
\multirow{4}{*}{\textbf{Accuracy}} & GBM & 0.5357($\pm$0.0698) & 0.7551($\pm$0.0602) & 0.5459($\pm$0.0697) & 0.6939($\pm$0.0645)\\
& LRM & 1($\pm$0.0000) &  1($\pm$0.0000) & 0($\pm$0.0000) & 0($\pm$0.0000)\\
& KNN & 1($\pm$0.0000) &  1($\pm$0.0000) & 1($\pm$0.0000) & 1($\pm$0.0000)\\
& SVM & 1($\pm$0.0000) &  1($\pm$0.0000) & 0($\pm$0.0000) & 0($\pm$0.0000)\\
\bottomrule
\end{tabular}
\caption{Classification accuracy of 32 synthetic-data-trained classifiers for larger COVID-19 incidence data of 18 countries or regions.}
\label{18CovidResult_Larger}
\end{table}

\begin{table}[H]
\centering
\begin{tabular}{lcrrrr}
\toprule
\multirow{2}{*}{\textbf{Evaluation}} & \multirow{2}{*}{\textbf{Model}} & \multicolumn{4}{c}{\textbf{22SF}}\\
\cmidrule(r){3-6}
& & \makecell[c]{WhiteN} & \makecell[c]{EnvN} & \makecell[c]{DemN}& \makecell[c]{MixedN} \\
\midrule
\multirow{4}{*}{\textbf{Accuracy}} & GBM &  0.8947($\pm$0.1380) &  0.5789($\pm$0.2220) &  0.0526($\pm$0.1004) &  0.8947($\pm$0.1380) \\
& LRM &  0($\pm$0.0000) &  0($\pm$0.0000) &  0.2105($\pm$0.1833) & 0.1053($\pm$0.1380) \\
& KNN &  1($\pm$0.0000) &  1($\pm$0.0000) & 1($\pm$0.0000) & 1($\pm$0.0000) \\
& SVM &  0.0526($\pm$0.1004) &  0($\pm$0.0000) &  0.3158($\pm$0.2090) &  0.1053($\pm$0.1380) \\
\midrule
\multirow{2}{*}{\textbf{Evaluation}} & \multirow{2}{*}{\textbf{Model}} &  \multicolumn{4}{c}{\textbf{5EWSI}}\\
\cmidrule(r){3-6} 
& & \makecell[c]{WhiteN} & \makecell[c]{EnvN} & \makecell[c]{DemN}& \makecell[c]{MixedN} \\
\midrule
\multirow{4}{*}{\textbf{Accuracy}} & GBM & 1($\pm$0.0000) &  0.9474($\pm$0.1004) &  0.0526($\pm$0.1004) &  0.9474($\pm$0.1004) \\
& LRM & 1($\pm$0.0000) & 1($\pm$0.0000) & 0($\pm$0.0000) & 0.1053($\pm$0.1351)\\
& KNN & 1($\pm$0.0000) & 1($\pm$0.0000) & 1($\pm$0.0000) & 1($\pm$0.0000)\\
& SVM & 1($\pm$0.0000) & 1($\pm$0.0000) & 0($\pm$0.0000) & 0.1053($\pm$0.1351)\\
\bottomrule
\end{tabular}
\caption{Classification accuracy of 32 synthetic-data-trained classifiers for larger COVID-19 Incidence Data of Singapore.}
\label{Singapore_Larger}
\end{table}

\newpage
\bibstyle{unsrt}
\bibliography{3rre}

\begin{thebibliography}{10}
\expandafter\ifx\csname url\endcsname\relax
  \def\url#1{\texttt{#1}}\fi
\expandafter\ifx\csname urlprefix\endcsname\relax\def\urlprefix{URL }\fi
\expandafter\ifx\csname href\endcsname\relax
  \def\href#1#2{#2} \def\path#1{#1}\fi

\bibitem{jones2020history}
D.~S. Jones, History in a crisis—lessons for covid-19, New England journal of medicine 382~(18) (2020) 1681--1683.

\bibitem{anderson2020will}
R.~M. Anderson, H.~Heesterbeek, D.~Klinkenberg, T.~D. Hollingsworth, How will country-based mitigation measures influence the course of the covid-19 epidemic?, The lancet 395~(10228) (2020) 931--934.

\bibitem{gargoum2021limiting}
S.~A. Gargoum, A.~S. Gargoum, Limiting mobility during covid-19, when and to what level? an international comparative study using change point analysis, Journal of Transport \& Health 20 (2021) 101019.

\bibitem{de2020initial}
Y.~B. De~Bruin, A.-S. Lequarre, J.~McCourt, P.~Clevestig, F.~Pigazzani, M.~Z. Jeddi, C.~Colosio, M.~Goulart, Initial impacts of global risk mitigation measures taken during the combatting of the covid-19 pandemic, Safety science 128 (2020) 104773.

\bibitem{tian2020investigation}
H.~Tian, Y.~Liu, Y.~Li, C.-H. Wu, B.~Chen, M.~U. Kraemer, B.~Li, J.~Cai, B.~Xu, Q.~Yang, et~al., An investigation of transmission control measures during the first 50 days of the covid-19 epidemic in china, Science 368~(6491) (2020) 638--642.

\bibitem{evans2024effectiveness}
S.~Evans, N.~R. Naylor, T.~Fowler, S.~Hopkins, J.~Robotham, The effectiveness and efficiency of asymptomatic sars-cov-2 testing strategies for patient and healthcare workers within acute nhs hospitals during an omicron-like period, BMC Infectious Diseases 24~(1) (2024) 64.

\bibitem{may1977thresholds}
R.~M. May, Thresholds and breakpoints in ecosystems with a multiplicity of stable states, Nature 269~(5628) (1977) 471--477.

\bibitem{scheffer2020critical}
M.~Scheffer, Critical transitions in nature and society, Vol.~16, Princeton University Press, 2020.

\bibitem{scheffer2001catastrophic}
M.~Scheffer, S.~Carpenter, J.~A. Foley, C.~Folke, B.~Walker, Catastrophic shifts in ecosystems, Nature 413~(6856) (2001) 591--596.

\bibitem{kefi2007spatial}
S.~K{\'e}fi, M.~Rietkerk, C.~L. Alados, Y.~Pueyo, V.~P. Papanastasis, A.~ElAich, P.~C. De~Ruiter, Spatial vegetation patterns and imminent desertification in mediterranean arid ecosystems, Nature 449~(7159) (2007) 213--217.

\bibitem{heffernan2005perspectives}
J.~M. Heffernan, R.~J. Smith, L.~M. Wahl, Perspectives on the basic reproductive ratio, Journal of the Royal Society Interface 2~(4) (2005) 281--293.

\bibitem{metcalf2017opportunities}
C.~J.~E. Metcalf, J.~Lessler, Opportunities and challenges in modeling emerging infectious diseases, Science 357~(6347) (2017) 149--152.

\bibitem{van2007slow}
E.~H. Van~Nes, M.~Scheffer, Slow recovery from perturbations as a generic indicator of a nearby catastrophic shift, The American Naturalist 169~(6) (2007) 738--747.

\bibitem{carpenter2006rising}
S.~R. Carpenter, W.~A. Brock, Rising variance: a leading indicator of ecological transition, Ecology letters 9~(3) (2006) 311--318.

\bibitem{dakos2010spatial}
V.~Dakos, E.~H. van Nes, R.~Donangelo, H.~Fort, M.~Scheffer, Spatial correlation as leading indicator of catastrophic shifts, Theoretical Ecology 3 (2010) 163--174.

\bibitem{guttal2008changing}
V.~Guttal, C.~Jayaprakash, Changing skewness: an early warning signal of regime shifts in ecosystems, Ecology letters 11~(5) (2008) 450--460.

\bibitem{biggs2009turning}
R.~Biggs, S.~R. Carpenter, W.~A. Brock, Turning back from the brink: detecting an impending regime shift in time to avert it, Proceedings of the National academy of Sciences 106~(3) (2009) 826--831.

\bibitem{tredennick2022anticipating}
A.~T. Tredennick, E.~B. O’Dea, M.~J. Ferrari, A.~W. Park, P.~Rohani, J.~M. Drake, Anticipating infectious disease re-emergence and elimination: a test of early warning signals using empirically based models, Journal of the Royal Society Interface 19~(193) (2022) 20220123.

\bibitem{harris2020early}
M.~J. Harris, S.~I. Hay, J.~M. Drake, Early warning signals of malaria resurgence in kericho, kenya, Biology letters 16~(3) (2020) 20190713.

\bibitem{o2021early}
D.~A. O'brien, C.~F. Clements, Early warning signal reliability varies with covid-19 waves, Biology Letters 17~(12) (2021) 20210487.

\bibitem{hastings2010regime}
A.~Hastings, D.~B. Wysham, Regime shifts in ecological systems can occur with no warning, Ecology letters 13~(4) (2010) 464--472.

\bibitem{dakos2015resilience}
V.~Dakos, S.~R. Carpenter, E.~H. van Nes, M.~Scheffer, Resilience indicators: prospects and limitations for early warnings of regime shifts, Philosophical Transactions of the Royal Society B: Biological Sciences 370~(1659) (2015) 20130263.

\bibitem{bury2021deep}
T.~M. Bury, R.~Sujith, I.~Pavithran, M.~Scheffer, T.~M. Lenton, M.~Anand, C.~T. Bauch, Deep learning for early warning signals of tipping points, Proceedings of the National Academy of Sciences 118~(39) (2021) e2106140118.

\bibitem{kong2021machine}
L.-W. Kong, H.-W. Fan, C.~Grebogi, Y.-C. Lai, Machine learning prediction of critical transition and system collapse, Physical Review Research 3~(1) (2021) 013090.

\bibitem{deb2022machine}
S.~Deb, S.~Sidheekh, C.~F. Clements, N.~C. Krishnan, P.~S. Dutta, Machine learning methods trained on simple models can predict critical transitions in complex natural systems, Royal Society Open Science 9~(2) (2022) 211475.

\bibitem{dakos2018identifying}
V.~Dakos, Identifying best-indicator species for abrupt transitions in multispecies communities, Ecological indicators 94 (2018) 494--502.

\bibitem{boers2021observation}
N.~Boers, Observation-based early-warning signals for a collapse of the atlantic meridional overturning circulation, Nature Climate Change 11~(8) (2021) 680--688.

\bibitem{kermack1927contribution}
W.~O. Kermack, A.~G. McKendrick, A contribution to the mathematical theory of epidemics, Proceedings of the royal society of london. Series A, Containing papers of a mathematical and physical character 115~(772) (1927) 700--721.

\bibitem{chakraborty2024early}
A.~K. Chakraborty, S.~Gao, R.~Miry, P.~Ramazi, R.~Greiner, M.~A. Lewis, H.~Wang, An early warning indicator trained on stochastic disease-spreading models with different noises (2024).
\newblock \href {http://arxiv.org/abs/2403.16233} {\path{arXiv:2403.16233}}.

\bibitem{lubba2019catch22}
C.~H. Lubba, S.~S. Sethi, P.~Knaute, S.~R. Schultz, B.~D. Fulcher, N.~S. Jones, catch22: Canonical time-series characteristics: Selected through highly comparative time-series analysis, Data Mining and Knowledge Discovery 33~(6) (2019) 1821--1852.

\bibitem{delong1988comparing}
E.~R. DeLong, D.~M. DeLong, D.~L. Clarke-Pearson, Comparing the areas under two or more correlated receiver operating characteristic curves: a nonparametric approach, Biometrics (1988) 837--845.

\bibitem{patrozou2009does}
E.~Patrozou, L.~A. Mermel, Does influenza transmission occur from asymptomatic infection or prior to symptom onset?, Public health reports 124~(2) (2009) 193--196.

\bibitem{dablander2022overlapping}
F.~Dablander, H.~Heesterbeek, D.~Borsboom, J.~M. Drake, Overlapping timescales obscure early warning signals of the second covid-19 wave, Proceedings of the Royal Society B 289~(1968) (2022) 20211809.

\bibitem{nanopoulos2001feature}
A.~Nanopoulos, R.~Alcock, Y.~Manolopoulos, Feature-based classification of time-series data, International Journal of Computer Research 10~(3) (2001) 49--61.

\bibitem{delecroix2023potential}
C.~Delecroix, E.~H. van Nes, I.~A. van~de Leemput, R.~Rotbarth, M.~Scheffer, Q.~Ten~Bosch, The potential of resilience indicators to anticipate infectious disease outbreaks, a systematic review and guide, PLOS Global Public Health 3~(10) (2023) e0002253.

\bibitem{southall2021early}
E.~Southall, T.~S. Brett, M.~J. Tildesley, L.~Dyson, Early warning signals of infectious disease transitions: a review, Journal of the Royal Society Interface 18~(182) (2021) 20210555.

\bibitem{gsell2016evaluating}
A.~S. Gsell, U.~Scharfenberger, D.~{\"O}zkundakci, A.~Walters, L.-A. Hansson, A.~B. Janssen, P.~N{\~o}ges, P.~C. Reid, D.~E. Schindler, E.~Van~Donk, et~al., Evaluating early-warning indicators of critical transitions in natural aquatic ecosystems, Proceedings of the National Academy of Sciences 113~(50) (2016) E8089--E8095.

\bibitem{thomson2001development}
M.~C. Thomson, S.~J. Connor, The development of malaria early warning systems for africa, Trends in parasitology 17~(9) (2001) 438--445.

\bibitem{mori2017early}
U.~Mori, A.~Mendiburu, S.~Dasgupta, J.~A. Lozano, Early classification of time series by simultaneously optimizing the accuracy and earliness, IEEE transactions on neural networks and learning systems 29~(10) (2017) 4569--4578.

\bibitem{hajian2013receiver}
K.~Hajian-Tilaki, Receiver operating characteristic (roc) curve analysis for medical diagnostic test evaluation, Caspian journal of internal medicine 4~(2) (2013) 627.

\bibitem{kuhn2008building}
M.~Kuhn, Building predictive models in r using the caret package, Journal of statistical software 28 (2008) 1--26.

\bibitem{dimitriadou2008misc}
E.~Dimitriadou, K.~Hornik, F.~Leisch, D.~Meyer, A.~Weingessel, Misc functions of the department of statistics (e1071), tu wien, R package 1 (2008) 5--24.

\bibitem{team2010r}
R.~D.~C. Team, R: A language and environment for statistical computing, (No Title) (2010).

\bibitem{robin2011proc}
X.~Robin, N.~Turck, A.~Hainard, N.~Tiberti, F.~Lisacek, J.-C. Sanchez, M.~M{\"u}ller, proc: an open-source package for r and s+ to analyze and compare roc curves, BMC bioinformatics 12~(1) (2011) 1--8.

\bibitem{fraser2009pandemic}
C.~Fraser, C.~A. Donnelly, S.~Cauchemez, W.~P. Hanage, M.~D. Van~Kerkhove, T.~D. Hollingsworth, J.~Griffin, R.~F. Baggaley, H.~E. Jenkins, E.~J. Lyons, et~al., Pandemic potential of a strain of influenza a (h1n1): early findings, science 324~(5934) (2009) 1557--1561.

\bibitem{zhong2003epidemiology}
N.~Zhong, B.~Zheng, Y.~Li, L.~Poon, Z.~Xie, K.~Chan, P.~Li, S.~Tan, Q.~Chang, J.~Xie, et~al., Epidemiology and cause of severe acute respiratory syndrome (sars) in guangdong, people's republic of china, in february, 2003, The Lancet 362~(9393) (2003) 1353--1358.

\bibitem{Government}
{Government Technology Agency of Singapore}, \href{https://www.tech.gov.sg/products-and-services/responding-to-covid-19-with-tech/}{Responding to covid-19 with tech} (2024).
\newline\urlprefix\url{https://www.tech.gov.sg/products-and-services/responding-to-covid-19-with-tech/}

\bibitem{cori2020package}
A.~Cori, S.~Cauchemez, N.~M. Ferguson, C.~Fraser, E.~Dahlqwist, P.~A. Demarsh, T.~Jombart, Z.~N. Kamvar, J.~Lessler, S.~Li, et~al., Package ‘epiestim’, CRAN: Vienna Austria (2020).

\bibitem{nash2023estimating}
R.~K. Nash, S.~Bhatt, A.~Cori, P.~Nouvellet, Estimating the epidemic reproduction number from temporally aggregated incidence data: A statistical modelling approach and software tool, PLoS Computational Biology 19~(8) (2023) e1011439.

\bibitem{SARS2003}
{Devakumar K. P.}, \href{https://www.kaggle.com/datasets/imdevskp/sars-outbreak-2003-complete-dataset?resource=download}{Sars 2003 outbreak dataset} (2020).
\newline\urlprefix\url{https://www.kaggle.com/datasets/imdevskp/sars-outbreak-2003-complete-dataset?resource=download}

\bibitem{lipsitch2003transmission}
M.~Lipsitch, T.~Cohen, B.~Cooper, J.~M. Robins, S.~Ma, L.~James, G.~Gopalakrishna, S.~K. Chew, C.~C. Tan, M.~H. Samore, et~al., Transmission dynamics and control of severe acute respiratory syndrome, science 300~(5627) (2003) 1966--1970.

\bibitem{liao2005clustering}
T.~W. Liao, Clustering of time series data—a survey, Pattern recognition 38~(11) (2005) 1857--1874.

\bibitem{gupta2020approaches}
A.~Gupta, H.~P. Gupta, B.~Biswas, T.~Dutta, Approaches and applications of early classification of time series: A review, IEEE Transactions on Artificial Intelligence 1~(1) (2020) 47--61.

\bibitem{Gao_2024_10967222}
S.~Gao, \href{https://zenodo.org/records/10967222}{{Code and Data for Early detection of disease outbreaks and non-outbreaks}}, zenodo, doi: 10.5281/zenodo.10967222, \url{https://zenodo.org/records/10967222} (Apr. 2024).
\newblock \href {https://doi.org/10.5281/zenodo.10967222} {\path{doi:10.5281/zenodo.10967222}}.
\newline\urlprefix\url{https://zenodo.org/records/10967222}

\bibitem{owidcoronavirus}
E.~Mathieu, H.~Ritchie, L.~Rodés-Guirao, C.~Appel, C.~Giattino, J.~Hasell, B.~Macdonald, S.~Dattani, D.~Beltekian, E.~Ortiz-Ospina, M.~Roser, Coronavirus pandemic (covid-19), Our World in DataHttps://ourworldindata.org/coronavirus (2020).

\bibitem{mietus2002pnnx}
J.~Mietus, C.~Peng, I.~Henry, R.~Goldsmith, A.~Goldberger, The pnnx files: re-examining a widely used heart rate variability measure, Heart 88~(4) (2002) 378--380.

\bibitem{wang2007structure}
X.~Wang, A.~Wirth, L.~Wang, Structure-based statistical features and multivariate time series clustering, in: Seventh IEEE international conference on data mining (ICDM 2007), IEEE, 2007, pp. 351--360.

\bibitem{held2004detection}
H.~Held, T.~Kleinen, Detection of climate system bifurcations by degenerate fingerprinting, Geophysical Research Letters 31~(23) (2004).

\bibitem{huisman2022estimation}
J.~S. Huisman, J.~Scire, D.~C. Angst, J.~Li, R.~A. Neher, M.~H. Maathuis, S.~Bonhoeffer, T.~Stadler, Estimation and worldwide monitoring of the effective reproductive number of sars-cov-2, Elife 11 (2022) e71345.

\bibitem{raghunath2020prediction}
S.~Raghunath, A.~E. Ulloa~Cerna, L.~Jing, D.~P. VanMaanen, J.~Stough, D.~N. Hartzel, J.~B. Leader, H.~L. Kirchner, M.~C. Stumpe, A.~Hafez, et~al., Prediction of mortality from 12-lead electrocardiogram voltage data using a deep neural network, Nature medicine 26~(6) (2020) 886--891.

\end{thebibliography}

\end{document}